\documentclass[journal]{IEEEtran}
\ifCLASSINFOpdf
\else
\fi
\usepackage{amsmath}
\usepackage{orcidlink}
\usepackage{amssymb}
\usepackage{amsthm}
\usepackage{algorithm}
\usepackage{algorithmic}
\usepackage{bbm}
\usepackage{enumitem}
\usepackage{multirow}
\usepackage{array}
\usepackage{graphicx}
\usepackage{placeins}

\begin{document}

\title{Boundary-Centric Clip-Budgeted Active Learning for Temporal Action Segmentation}

\author{
\IEEEauthorblockN{Halil Ismail Helvaci\textsuperscript{\,1}\,\orcidlink{0000-0003-0024-6943}} \hspace{0.5cm}
\IEEEauthorblockN{Sen-ching Samson Cheung\textsuperscript{\,1}\,\orcidlink{0000-0002-9207-5514}} \\
\vspace{0.5cm}
\IEEEauthorblockA{\textit{\textsuperscript{\,1}Department of Electrical and Computer Engineering, University of Kentucky, Lexington, KY, USA}} \\
\and

%
}


\maketitle

\begin{abstract}

Temporal action segmentation (TAS) in untrimmed videos requires dense temporal supervision. However, most of the annotation cost is spent identifying action transitions where segmentation errors concentrate and small temporal shifts can disproportionately degrade segment-level metrics. We introduce B-ACT, a clip-budgeted active learning framework that explicitly allocates supervision to these error-prone boundary regions. B-ACT operates in a hierarchical two-stage loop: (i) it ranks and queries unlabeled videos using predictive uncertainty, and (ii) within each selected video, it detects candidate transitions from the current model predictions and selects the top-$K$ boundaries via a novel boundary score. The boundary score fuses neighborhood uncertainty, class ambiguity, and temporal prediction dynamics to reveal the underlying importance of each frame. Importantly, our annotation protocol requests labels only at the boundary frames while still training on boundary-centered clips to exploit temporal context through the model's receptive field. Extensive experiments on GTEA, 50Salads, and Breakfast demonstrate that boundary-centric supervision delivers strong label efficiency and consistently surpasses representative TAS active learning baselines and prior state of the art under sparse budgets. Gains are largest on datasets where performance is highly sensitive to boundary placement, as measured by edit and overlap-based F1 metrics.
\end{abstract}


\IEEEpeerreviewmaketitle

\section{Introduction}
\label{sec:introduction}

Understanding human activities automatically in long, unstructured videos is essential for a wide range of applications, including robotics, healthcare, and daily activity recognition \cite{helvaci2024localizing, manakitsa2024review, helvaci2025hrtr, ding2023temporal}. This has led to extensive research on temporal action segmentation (TAS), which aims to assign an action label to each frame and partition a long video into consecutive non-overlapping segments.

In principle, TAS requires dense temporal supervision—provided either as frame-level labels or segment-level annotations with precise boundary localization \cite{ding2023temporal}. This process is costly, as untrimmed videos are often long, requiring annotators to review hours of footage and precisely mark action boundaries throughout each video, making data collection difficult to scale. To reduce annotation cost, prior studies on TAS have investigated semi-supervised approaches that leverage a limited set of fully annotated videos alongside a larger collection of unlabeled videos \cite{singhania2022iterative, ding2022leveraging, wang2021self}. Nevertheless, these methods still require full annotations on multiple untrimmed videos, and do not address how to identify the most informative videos for annotation and subsequent training. Weakly supervised methods instead rely on coarse annotations for all training videos \cite{chang2019d3tw, chang2021learning, ding2018weakly, huang2016connectionist, kuehne2017weakly}, but their performance remains significantly below fully supervised approaches. 

Time-stamp annotation \cite{khan2022timestamp, li2021temporal, liu2023reducing, rahaman2022generalized} achieves performance closer to fully-supervised methods by using only a small number of labeled frames per video. However, obtaining these timestamps remains costly, as annotators must still watch every video in its entirety and manually select frames corresponding to distinct actions. A central limitation of these methods is the absence of principled strategies for selecting which videos and frames/clips to label, which could substantially reduce annotation time using a more efficient, guided labeling process.

Another strategy for reducing annotation cost is the conventional active learning (AL) paradigm, which iteratively selects unlabeled samples for labeling according to a utility criterion \cite{ren2021survey}. Nonetheless, most existing AL methods focus on sample-level classification, where each instance is associated with a single label. This setting differs fundamentally from TAS, in which each video comprises multiple temporally dependent labels, requiring both video-level and frame-level selection. Recent hybrid AL for action detection \cite{rana2022all, rana2023hybrid, joudaki2025new} focus on annotating short spatiotemporal regions, but are limited to short single-action clips. In contrast, TAS entails long untrimmed videos with multiple temporally-dependent actions and substantial background content.

A key challenge in applying AL to TAS is that errors are not uniformly distributed over frames, but concentrate around temporal transitions. Prior work consistently identifies boundary ambiguity and over-segmentation as primary sources of error, motivating architectures and losses that directly target boundaries (boundary regression, and boundary-smoothing refinements) rather than improving “interior” frames alone \cite{wang2020boundary, wang2023dir, chen2022uncertainty, wang2024faster, ding2023temporal, farha2019ms}. These methods also emphasize that temporal continuity should only change at true boundaries, reinforcing the importance of accurate boundary localization \cite{ding2023temporal}. Moreover, uncertainty modeling in TAS consistently highlights transition regions as the most ambiguous and informative parts of the sequence \cite{chen2022uncertainty}. This suggests that annotation and learning should be concentrated on boundary regions. In this work, we propose a boundary-centric clip-budgeted active learning framework that explicitly targets these high-impact regions to maximize label efficiency.

\noindent\textbf{Key Contributions.} We propose B-ACT (Boundary-centric Active Learning for TAS), a clip-budgeted active learning framework that concentrates supervision on action boundaries, where segmentation errors are most pronounced. Our main contributions include: 

\begin{itemize}
    \item A clip-budgeted annotation protocol that queries only $K$ boundary frames per selected video while leveraging $\ell$-frame temporal context around each queried boundary.

    \item An uncertainty-guided acquisition policy that selects informative unlabeled videos and prioritizes candidate action transitions using a novel boundary score capturing local confusion, classification ambiguity, and temporal prediction change.

    \item An efficient two-stage AL loop for TAS that substantially reduces labeling compared to full-video annotation.
\end{itemize}

Extensive experiments on GTEA \cite{fathi2011learning}, Breakfast \cite{kuehne2014breakfast}, and 50Salads \cite{stein2013combining} demonstrate consistent gains over prior AL baselines and state-of-the-art.

\begin{figure}[h]
    \centering
    \includegraphics[width= 1.0\linewidth]{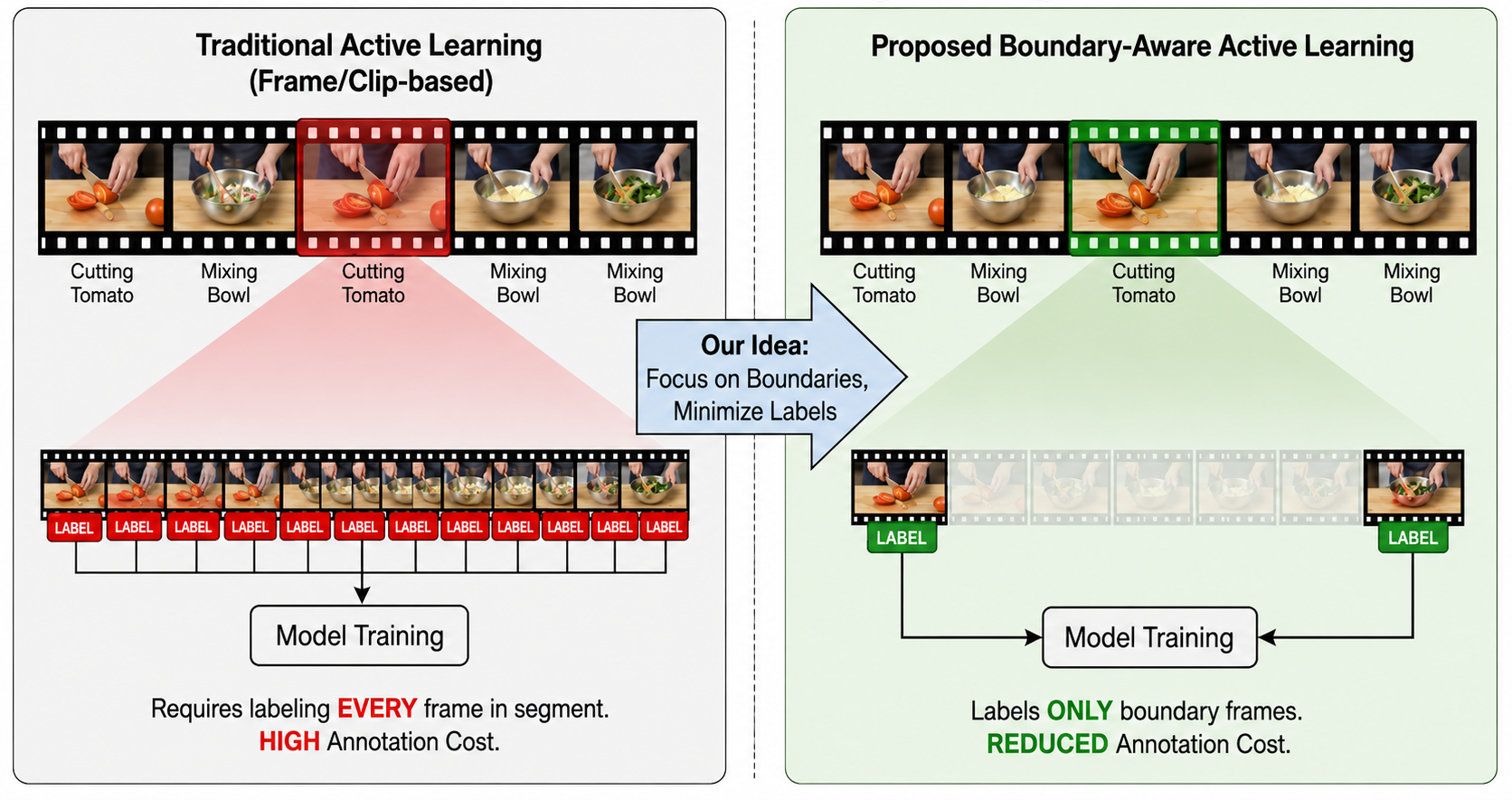}
    \caption{Comparison of active learning strategies for video. (a) Traditional frame or clip based active learning queries samples within action segments and often requires dense frame-level annotation, resulting in high labeling effort. (b) Proposed boundary-aware active learning prioritizes action boundary frames for querying, requiring labels only at segment transitions and thus reducing annotation cost while still supporting effective model training.}
    \label{fig:predictions}
\end{figure}

\section{Related Work}
\label{sec:related_work}

\textbf{Temporal action segmentation.} TAS has been extensively studied under full supervision, including recurrent models \cite{richard2017weakly, singh2016multi}, temporal convolutional networks (TCNs) \cite{farha2019ms, lea2017temporal}, transformers \cite{yi2021asformer, zhang2022actionformer}, graph neural networks \cite{huang2020improving, zhang2022semantic2graph}, diffusion models \cite{liu2023diffusion, gong2024actfusion} and hybrid designs that combine convolutional and attention-based backbones \cite{lu2024fact}. A dominant line of work such as MS-TCN \cite{farha2019ms}, ASRF \cite{ahn2021refining}, and ASFormer \cite{yi2021asformer}, focus on multi-stage refinement, where predictions are iteratively refined to improve temporal consistency and reduce over-segmentation.

To reduce annotation cost, semi-supervised TAS leverages a subset of densely annotated videos (frame-level) \cite{singhania2022iterative, ding2022leveraging,wang2021self}, while weakly-supervised TAS relies on coarse supervision such as action transcripts/action sets \cite{chang2019d3tw, chang2021learning, ding2018weakly, huang2016connectionist, kuehne2017weakly, li2019weakly}, sparse frame labels \cite{bueno2023leveraging, khan2022timestamp, li2021temporal} or video-level labels \cite{ding2022leveraging, ding2023temporal}. Timestamp-based supervision reduces labeling density by training from a small set of annotated frames per action \cite{khan2022timestamp,li2021temporal,liu2023reducing,rahaman2022generalized}, but still typically requires annotators to review entire videos. In contrast, our work focuses on actively selecting both video and short temporal regions within each video for annotation, under a fixed budget.

Unsupervised and self-supervised pretraining has also been explored to reduce reliance on dense labels, including learning representations from large-scale instructional video corpora \cite{miech2019howto100m} and transferring to TAS \cite{aakur2019perceptual, bansal2022my, dvornik2023stepformer, elhamifar2020self, kukleva2019unsupervised, sarfraz2021temporally, shah2023steps}. Rather than removing annotation entirely, we focus on making annotation more efficient by allocating supervision to the temporally most informative regions.

\textbf{Boundary-Aware Action Segmentation.} A central challenge in TAS is boundary ambiguity: predictions become unreliable in the transition region between consecutive actions, leading to boundary jitter and over-segmentation that disproportionately degrades segment-level metrics \cite{wang2020boundary, wang2023dir, chen2022uncertainty, ding2023temporal}. This concentration of errors around temporal transitions is also a key reason TAS does not reduce cleanly to conventional sample-level AL, since the most informative supervision is not uniformly distributed over frames but is typically localized near boundaries \cite{chen2022uncertainty, ding2023temporal}. Consequently, many methods explicitly model boundaries by enforcing temporal continuity within segments and introducing architectures or losses that target boundary refinement \cite{farha2019ms, ding2023temporal}.

Explicitly, boundary-aware architectures improve predictions via targeted refinement: BCN uses boundary-conditioned cascades \cite{wang2020boundary}, ASRF predicts boundary probabilities to suppress fragments \cite{ahn2021refining}, and recent transformers like BaFormer use class-agnostic boundaries to guide segmentation \cite{wang2024baformer}. Chen et al. \cite{chen2022uncertainty} show that predictive ambiguity concentrates in transitions, motivating explicit uncertainty modeling. These works consistently identify boundaries as high-impact supervision targets. This observation motivates our boundary-centric acquisition strategy, which prioritizes labeling at uncertain transitions

\textbf{Uncertainty Estimation.} Uncertainty-based acquisition is a core paradigm in AL, where informative samples are selected using entropy, margin, and committee disagreement to select informative samples \cite{ren2021survey, beluch2018power}. In deep AL, epistemic uncertainty is commonly approximated using Monte Carlo dropout (MCD) \cite{gal2016dropout}, deep ensembles \cite{lakshminarayanan2017simple}, and other Bayesian surrogates \cite{kendall2017uncertainties}. These estimators are often paired with information-theoretic acquisition functions such as BALD \cite{houlsby2011bayesian}, and its variants \cite{gal2017deep}. For batch querying, uncertainty is frequently complemented by diversity objectives (e.g., core-set selection) to reduce redundancy \cite{sener2017active}.

For dense prediction, informativeness is typically localized, motivating region-level acquisition querying informative regions/super-pixels using uncertainty and inconsistency signals \cite{kasarla2019, siddiqui2020viewal, cai2021revisiting}. This aligns naturally with TAS, where uncertainty concentrates near temporal boundaries, requiring per-frame uncertainty estimation and aggregation into clip-level selection.

\textbf{Active Learning.} AL aims to reduce annotation effort by iteratively selecting the most informative unlabeled examples for labeling and retraining the model with newly acquired annotations \cite{ren2021survey}. Most modern approaches are guided by two complementary criteria: uncertainty and diversity. Uncertainty-based methods \cite{nguyen2022measure} prioritize samples of which the current model is least confident, whereas diversity-based methods \cite{sener2017active} aim to ensure the selected samples cover the data distribution and avoid redundant queries.

For TAS, Su \textit{et al.} \cite{su2024two} introduced a two-stage AL framework that measures informativeness through sequence alignment. In the \emph{inter-video} stage, unlabeled videos are prioritized based on high alignment cost to labeled prototypes, encouraging diversity in temporal structure and action ordering. In the \emph{intra-video} stage, clips are selected via an alignment-based summarization objective using dynamic time warping (DTW), including drop-DTW \cite{dvornik2021drop}, to promote sequential coverage within each queried video.

While alignment-based acquisition in \cite{su2024two} effectively captures global diversity in action ordering, it does not directly address the boundary-localized nature of TAS errors established in literature. As extensively documented in \cite{wang2020boundary, wang2023dir, chen2022uncertainty, ding2023temporal}, segmentation failures concentrate at ambiguous action transitions, where small temporal shifts disproportionately degrade segmental metrics such as edit score and F1 score. As a result, structural dissimilarity does not necessarily correlate with learning value. Compared with the labeled prototypes, a video may be globally diverse yet contain few informative boundaries, while a structurally similar video may contain critical uncertain transitions.  This observation motivates acquisition strategies that directly quantify model uncertainty at the frame level and allocate supervision to temporally critical boundary regions.

In this work, we adopt a two-stage paradigm of \cite{su2024two}, video selection followed by clip selection, but modify both stages to explicitly target boundary uncertainty. First, we select videos using Monte Carlo dropout-based uncertainty estimation \cite{gal2016dropout} rather than alignment-to-prototype cost, prioritizing videos with high predictive uncertainty. Second, within selected videos, we employ a boundary-centric scoring function that prioritizes clips around uncertain action transitions, rather than alignment-based summarization for sequential coverage. This approach significantly improves label efficiency by concentrating on boundary regions that dominate TAS error.

\section{Methodology}
\label{sec:methodology}

\textbf{Problem Setting.} Given an untrimmed input video represented by a sequence of pre-extracted frame features $\mathbf{X}=\{\mathbf{x}_t\}_{t=1}^{T}$ with $\mathbf{x}_t\in\mathbb{R}^{D}$, where $D$ is the feature dimension, TAS aims to predict frame-wise action labels $\mathbf{Y}=\{y_t\}_{t=1}^{T}$, where $y_t\in\{1,\ldots,C\}$ action classes. For a dataset of $N_{\text{vid}}$ videos, the $i$-th video is written as $\mathbf{X}^{(i)}=\{\mathbf{x}^{(i)}_t\}_{t=1}^{T_i}$, where $T_i$ denotes the number of frames in video $i$. In standard video AL, querying $\mathbf{X}^{(i)}$ incurs a labeling cost of $T_i$, since all $T_i$ frames must be annotated.

We propose a boundary-centric clip-budgeted AL protocol that allocates supervision to short temporal clips around predicted action transitions. Each AL round consists of two decisions: (i) video selection: selecting unlabeled videos to query, and (ii) clip selection: selecting boundary locations within those videos to annotate. Importantly, only boundary frames are labeled, while surrounding frames are used as unlabeled temporal context.

\begin{figure*}[h!]
    \centering
    \includegraphics[width= 0.8\linewidth]{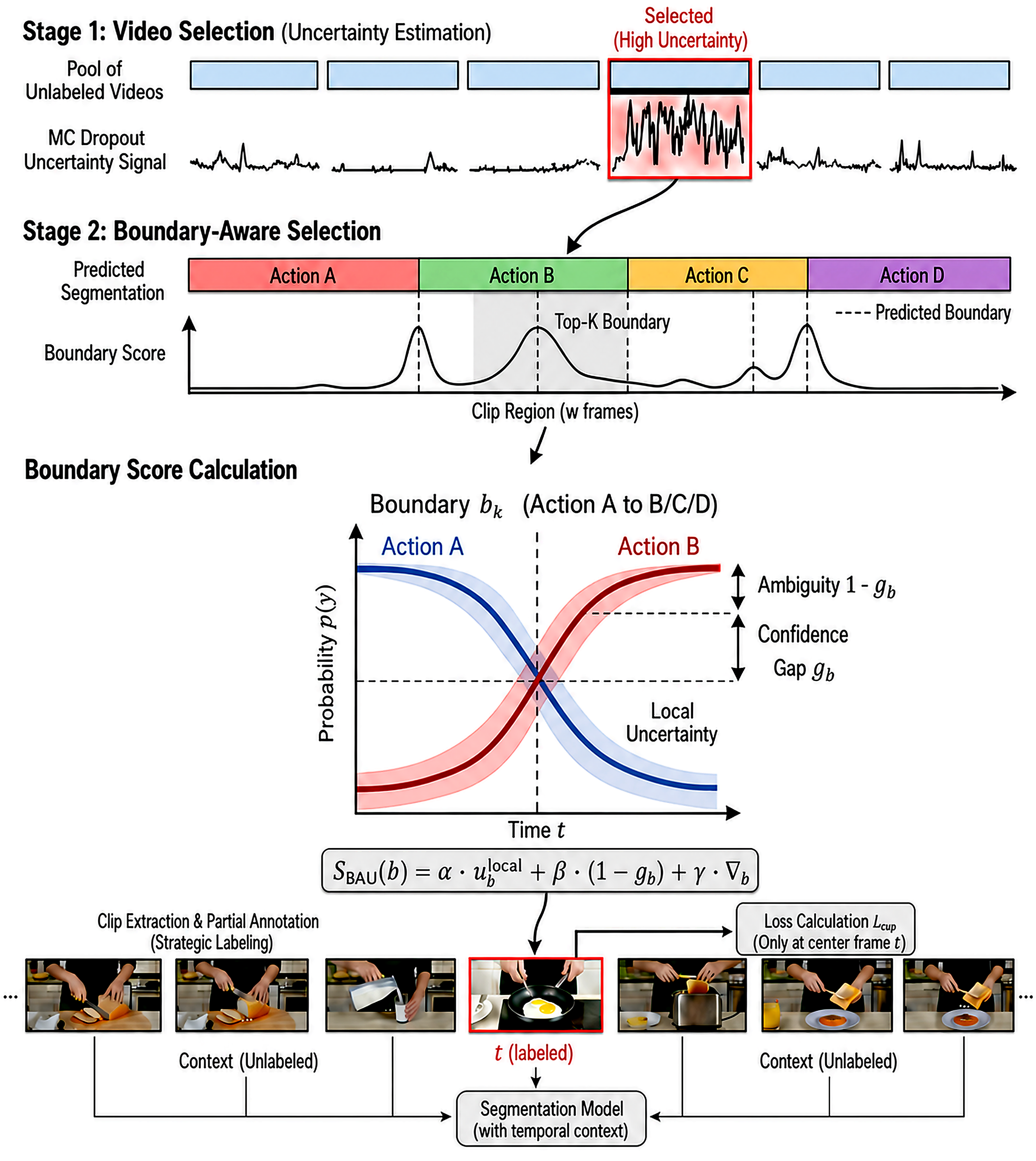}
    \caption{\textbf{Overview of the proposed boundary-centric active learning pipeline for TAS.} \textbf{Stage 1:} Unlabeled videos are ranked by predictive uncertainty estimated via MC dropout, and the most uncertain videos are selected. \textbf{Stage 2:} For each selected video, candidate boundaries are extracted from label changes in the predicted sequence and ranked by a boundary score that fuses (i) local predictive uncertainty in a temporal neighborhood, (ii) boundary ambiguity via the inverse top-1/top-2 confidence gap, and (iii) a temporal gradient measuring predictive distribution change across time. The top-$K$ boundaries define short clip regions for annotation; only the center boundary frame is labeled while surrounding frames provide temporal context during training.}

    \label{fig:actions}
\end{figure*}

\textbf{Framework Overview.} Our framework addresses two core challenges: identifying informative videos and selecting informative temporal regions within. The complete pipeline (Fig.~\ref{fig:actions}) follows a hierarchical two-stage selection. In Stage 1, we estimate predictive uncertainty for each unlabeled video using Monte Carlo Dropout \cite{gal2016dropout}, and select the top $N_q$ most uncertain ones for the next stage. In Stage 2, for each selected video, candidate action boundaries are identified and ranked using a boundary score that captures local uncertainty, class ambiguity, and temporal prediction dynamics. We then annotate clips centered at the top-$K$ boundaries, labeling only the boundary frame $b_k$ while using surrounding frames as unlabeled context during training.

This design concentrates supervision at high-information decision boundaries while enabling the model to learn smooth temporal predictions through the temporal receptive fields. The model exploits temporal smoothness by leveraging unlabeled frames between annotated boundaries to propagate supervision through its temporal receptive field. This annotation incurs a cost of $N_q \cdot K$ labels per iteration while providing training context of $N_q \cdot K \cdot \ell$ frames for training.

\subsection{Uncertainty-Guided Boundary-Centric Active Learning}
\label{sec:framework}



We propose a boundary-centric AL paradigm that partially annotates queried videos by concentrating labels near predicted action transitions. Given a queried video $\mathbf{X}^{(i)}$ of length $T_i$, we predict candidate temporal boundaries and select the top-$K$ boundary locations $\{b_k^{(i)}\}_{k=1}^{K}$, where $K$ is the number of queried boundaries per video and $b_k^{(i)} \in \{1,\ldots,T_i\}$ denotes the frame index of the $k$-th selected boundary in video $i$. Around each selected boundary, we define a boundary-centered temporal window

\begin{equation}
I_k^{(i)}=
\left[b_k^{(i)}-\left\lfloor \frac{\ell}{2} \right\rfloor,\;
      b_k^{(i)}+\left\lfloor \frac{\ell}{2} \right\rfloor\right]\cap[1,T_i]
\end{equation}
and extract the corresponding local clip
\begin{equation}
\mathcal{C}_k^{(i)}=\{\mathbf{x}_t^{(i)}\}_{t\in I_k^{(i)}}
\end{equation}
where $\ell$ denotes the nominal clip length in frames. Near the beginning or end of the video, $I_k^{(i)}$ is truncated by the intersection with $[1,T_i]$ to keep indices within range. We request supervision only at the selected boundary frame $y_{b_k^{(i)}}^{(i)}$ for each clip. Thus, each queried video incurs only $K$ frame labels, while exposing the model to up to $K\ell$ contextual frames during training. Since $K\ell\ll T_i$, the protocol substantially reduces annotation relative to dense video labeling.

For the active learning loop, the training pool is denoted by $\mathcal{D}_{\text{train}}=\{\mathbf{X}^{(i)}\}_{i=1}^{N_{\text{vid}}}$. At round $r \in \{1,\ldots,R\}$, the labeled and unlabeled subsets are denoted by $\mathcal{D}_L$ and $\mathcal{D}_U$, with $\mathcal{D}_L \cup \mathcal{D}_U=\mathcal{D}_{\text{train}}$ and $\mathcal{D}_L \cap \mathcal{D}_U=\emptyset$. We select a query batch $\mathcal{S}_{\mathrm{query}}^{(r)} \subset \mathcal{D}_U$ containing $N_q$ videos to be queried in each round. For each selected video, we predict candidate boundaries, annotate the top-$K$ selected boundary frames, and add the newly supervised samples to $D_L$. We repeat this process for $R$ rounds under a total labeled-frame budget of $B$

\begin{equation}
|\Omega_L|\le B
\end{equation}
where $\Omega_L$ is the set of frames labeled so far. After each round, the segmentation model
$M_\theta$ is retrained on $\mathcal{D}_L$.

The choice to annotate a single frame per clip—specifically the predicted boundary frame $b_k$—requires justification on two grounds: (i) why one label can provide useful supervision, and (ii) how the queried frame can be labeled reliably.

\textbf{Why boundary-frame supervision is effective.} Temporal action segmentation models rely heavily on temporal context: predictions at a given frame are informed not only by local visual evidence but also by neighboring observations across time. Consequently, supervision applied at a single frame can influence predictions beyond the labeled instant. As such, we place supervision at predicted action boundaries, which are known to concentrate segmentation ambiguity and localization errors. The surrounding unlabeled clip context enables the model to associate the queried transition with the preceding and subsequent temporal patterns. This design is also consistent with prior timestamp-supervision studies \cite{khan2022timestamp, liu2023reducing, rahaman2022generalized}, which show that sparse frame-level labels can support competitive temporal action segmentation when temporal structure is exploited effectively. Our setting, however, differs in two important respects: first, queried frames are selected actively rather than uniformly or manually; second, supervision is concentrated near predicted transitions and paired with boundary-centered temporal context during training.

\textbf{Practical labeling of boundary frames.} A key practical concern 
is whether an annotator can reliably assign an action label to a single frame without watching the surrounding video context. In our protocol, the annotation request is defined at a single boundary frame at $b_k$. In practice, the interface may display minimal neighboring context to help the annotator assign the correct action label, while the recorded supervision remains a single frame label. The surrounding $\ell$ context frames are not annotated and are used only by the model during training as unlabeled temporal context. The measured supervision cost therefore remains K labeled frames per queried video, although practical annotation time depends on how the annotator interfaces with the queried frame. The clip length $\ell$ is chosen (ablated in Tab.~\ref{tab:clip_length}) to maximize the model's temporal receptive field around each boundary, not to assist the annotator.

The feasibility of single-frame annotation at boundaries is supported  by the timestamp-supervision literature~\cite{khan2022timestamp,
farha2019ms,liu2023reducing}, which demonstrates that annotators can 
reliably assign action labels to individual frames when the action class vocabulary is well-defined. In our active selection setting, the boundary frame $b_k$ is specifically chosen because the model predicts a label change at that location, making it more likely to correspond to a semantically meaningful change than a uniformly sampled frame. To guard against misassignment near ambiguous transitions, the model's temporal context (unlabeled frames in the clip) provides implicit regularization: if the label at $b_k$ is inconsistent with the surrounding unlabeled predictions, the loss gradient will push the model toward a coherent temporal explanation, effectively mitigating minor annotation errors at boundaries.

\subsection{Uncertainty Quantification}

To guide active selection, we require a frame-level estimate of predictive uncertainty. We employ Monte Carlo Dropout (MCD), which offers a computationally efficient approximation to Bayesian uncertainty estimation \cite{gal2016dropout}. In the ideal Bayesian setting, one would evaluate the full posterior predictive distribution:
\begin{equation}
p(y_t \mid \mathbf{x}_t, \mathcal{D}_L) = \int p(y_t \mid \mathbf{x}_t, \theta) \, p(\theta \mid \mathcal{D}_L) \, d\theta
\label{eq:bayesian_posterior}
\end{equation}
where the integral marginalizes over all parameter configurations $\theta$, weighted by the posterior $p(\theta \mid \mathcal{D}_L)$ conditioned on the labeled training set. In practice, exact evaluation of Eq.~\eqref{eq:bayesian_posterior} is intractable for deep neural networks, so we approximate it via Monte Carlo estimation using dropout-induced stochasticity:

\begin{equation}
p(y_t \mid \mathbf{x}_t, \mathcal{D}_L) \approx \frac{1}{S} \sum_{s=1}^{S} p(y_t \mid \mathbf{x}_t, \theta_s)
\label{eq:mc_dropout}
\end{equation}
where $\theta_s$ denotes the network parameters under a dropout mask $s$ sampled at inference time. Keeping dropout active during inference and executing $S$ stochastic forward passes yields $S$ predictive samples. Let $\mathbf{p}_{t,s}\in\mathbb{R}^{C}$ denote the class-probability vector at frame $t$ for the $s$-th dropout sample, and define the mean prediction as $\bar{\mathbf{p}}_t=\frac{1}{S}\sum_{s=1}^{S}\mathbf{p}_{t,s}$. We quantify frame-level uncertainty using predictive entropy:
\begin{equation}
u_t = H[\bar{\mathbf{p}}_t] = -\sum_{c=1}^C \bar{p}_{t,c} \log \bar{p}_{t,c}
\label{eq:frame_uncertainty}
\end{equation}

High entropy indicates the model is uncertain which action class applies at frame $t$, while low entropy reflects confident predictions. This frame-level uncertainty forms the foundation for both video selection and boundary scoring.

\subsection{Video Selection}
\label{sec:video_selection}

For each unlabeled video $\mathbf{X}^{(j)} \in \mathcal{D}_U$, we estimate frame-level uncertainties
$\{u_t^{(j)}\}_{t=1}^{T_j}$ via Eq.~\eqref{eq:frame_uncertainty} with $S$ MCD samples, and aggregate them into a video-level informativeness score via mean pooling: 

\begin{equation}
U^{(j)} = \frac{1}{T_j} \sum_{t=1}^{T_j} u_t^{(j)}
\end{equation}
Videos with high average uncertainty $U^{(j)}$ contain broader regions of predictive ambiguity and are therefore more likely to benefit from additional supervision. Since boundary localization is performed separately in Stage 2, the video-level score is intended to identify globally informative videos with mean pooling rather than individual transition points. We therefore select the top-$N_q$ videos with the highest uncertainty:
\begin{equation}
\mathcal{S}_{\text{query}} =
\operatorname*{arg\,max}_{\mathcal{S}\subset \mathcal{D}_U,\ |\mathcal{S}|=N_q}\ \sum_{\mathbf{X}^{(j)}\in \mathcal{S}} U^{(j)}
\end{equation}
This uncertainty-driven criterion explicitly prioritizes videos where the model exhibits the greatest predictive ambiguity, and are therefore expected to yield the greatest benefit from additional supervision.

\subsection{Boundary-Centric Clip Selection}
\label{sec:clip_selection}

Having identified informative videos, we next localize the temporal regions that are most valuable to annotate. Prior work suggests that temporal segmentation errors are disproportionately concentrated around action boundaries \cite{wang2020boundary, chen2022uncertainty, ding2023temporal}. We therefore formulate clip selection to explicitly focus on transitions. For each queried video $\mathbf{X}^{(j)} \in \mathcal{S}_{\text{query}}$, we first extract predicted action boundaries from the model outputs. Let $\hat{y}_t^{(j)} = \operatorname*{arg\,max}_{c} \ \bar{p}_{t,c}^{(j)}$ denote the predicted label at frame $t$ obtained over maxing the mean MCD probability vector over different action class $c$. We then define the set of boundary indices as frames where the predicted label changes between adjacent timesteps:
\begin{equation}
\mathcal{B}_j = \left\{ t \in \{2,\ldots,T_j\} \;:\; \hat{y}_t^{(j)} \neq \hat{y}_{t-1}^{(j)} \right\}
\end{equation}
This boundary extraction yields a set of candidate transition frames that serve as anchors for downstream clip selection and annotation.

\textbf{Motivation for the boundary score.} A principled acquisition function for boundary regions should capture three qualitatively distinct notions of difficulty, each supported by prior TAS literature. First, a boundary is more informative if the model is already uncertain in its immediate temporal vicinity: Chen et al. [23] demonstrate that predictive ambiguity concentrates in the transition neighborhood, not only at the frame of the label change itself, making neighborhood-level uncertainty a robust proxy for annotation value. Second, a boundary is harder to disambiguate if the two most likely classes are nearly tied at the transition frame: this concept of margin-based ambiguity is a canonical signal in active learning theory [17], [52], and it captures decision fragility that global entropy may understate when the competing classes form a dominant pair. Third, a boundary is more genuine—and more critically in need of supervision—if the model's predicted distribution shifts sharply across it: BCN [21] and ASRF [39] both highlight that true transitions are characterized by abrupt changes in action evidence, whereas spurious over-segmentation fragments produce small, gradual distributional drifts. Targeting boundaries with large temporal gradients therefore simultaneously promotes label efficiency and directly counteracts the over-segmentation failure mode that these architectures were designed to address. We operationalize these three motivations as complementary, computationally lightweight signals.

For each predicted boundary $b \in \mathcal{B}_j$, we define a symmetric window $W_b = [b-w, b+w]\cap[1,T_j]$ with $w=\left\lfloor \ell/2 \right\rfloor$. Within $W_b$, we compute three complementary uncertainty signals.

\textbf{Local uncertainty.} We first characterize the average predictive uncertainty in the neighborhood of a candidate boundary. The motivation is that, in TAS, most failures concentrate around action transitions, where visual evidence is ambiguous and temporal context is critical. We therefore aggregate frame-level uncertainty within a local window $W_b$ centered at boundary index $b$, yielding a robust estimate of uncertainty surrounding the candidate transition. This term promotes querying boundaries that are surrounded by consistently high uncertainty, which are expected to offer high annotation utility for improving boundary localization and reducing over-segmentation:
\begin{equation}
u_{b}^{\text{local},(j)} = \frac{1}{|W_b|} \sum_{t \in W_b} u_t^{(j)} .
\end{equation}

\textbf{Confidence gap.} We next quantify class-level ambiguity at boundary frames using the margin between the most probable (top-1) and second most probable (top-2) predicted classes. Margin-based query selection is a theoretically grounded acquisition strategy in AL \cite{ren2021survey, beluch2018power}, as a small margin implies that the decision boundary lies close to the corresponding sample, where additional label information most effectively reduces classification risk. In TAS, this consideration is particularly salient at action transitions, where predictive uncertainty often concentrates between competing neighboring actions. Consequently, the top-2 margin provides a more precise indicator than full-distribution entropy for identifying unstable boundary predictions. A smaller margin reflects greater predictive ambiguity and therefore a higher priority for annotation:


\begin{equation}
g_b^{(j)} = p_{b}^{\max,(j)} - p_{b}^{2\text{nd},(j)}
\end{equation}
where $p_{b}^{\max,(j)}=\max_c \bar{p}_{b,c}^{(j)}$ and $p_{b}^{2\text{nd},(j)}$ is the second-largest value in $\{\bar{p}_{b,c}^{(j)}\}_{c=1}^{C}$. 


\textbf{Temporal gradient.} As the last signal, we quantify the sharpness of distributional change across the candidate boundary. Boundary-aware architectures such as BCN \cite{wang2020boundary} and ASRF \cite{ahn2021refining} are explicitly designed around the observation that genuine action transitions coincide with abrupt shifts in visual evidence, whereas over-segmentation artifacts are produced by gradual, low-confidence drifts. A large frame-to-frame L2 distance between consecutive predicted distributions therefore indicates a boundary that is both temporally sharp and likely to constitute a true transition, making it a high-value annotation target. Conversely, boundaries with small temporal gradients are more likely to be spurious fragments whose supervision benefit is limited. We define this in practical terms as follows:

\begin{equation}
\nabla_b^{(j)} = \frac{1}{|E_b|} \sum_{t \in E_b} \left\|\bar{\mathbf{p}}_{t+1}^{(j)} - \bar{\mathbf{p}}_{t}^{(j)}\right\|_2
\end{equation}
where $E_b=\{t \mid t \in W_b,\; t+1 \in W_b\}$ is the set of valid adjacent indices fully contained in $W_b$ (i.e., the time steps for which the pair $(t,t+1)$ exists inside the window), and $|E_b|$ normalizes by the number of such pairs when $W_b$ is truncated at the sequence boundaries.


\textbf{Boundary Score.} We integrate the three signals into a unified boundary informativeness score:

\begin{equation}
S_{\mathrm{BAU}}^{(j)}(b) = \alpha \cdot u_{b}^{\text{local},(j)} + \beta \cdot (1 - g_b^{(j)}) + \gamma \cdot \nabla_b^{(j)}.
\label{eq:boundary_score}
\end{equation}
The three terms correspond to the three complementary failure modes identified above: neighborhood confusion (local uncertainty), decision fragility (inverted margin), and transition sharpness (temporal gradient). We invert the confidence gap via $(1-g_b^{(j)})$ so that smaller top-1/top-2 margins—indicating greater ambiguity—receive higher scores. The weights $\alpha$, $\beta$, $\gamma$ control the relative emphasis on each signal. We tune these weights via grid search on a held-out validation split (see Section \ref{sec:implementation_details}  for details), yielding $\alpha=0.2$, $\beta=0.3$, and $\gamma=0.5$.

Given the boundary scores, we select the top-$K$ candidates via a cardinality-constrained maximization:
\begin{equation}
\{b_1^{(j)}, \ldots, b_K^{(j)}\} =
\operatorname*{arg\,max}_{B \subset \mathcal{B}_j,\ |B|=K}\ \sum_{b \in B} S_{\mathrm{BAU}}^{(j)}(b).
\end{equation}
For each selected boundary $b_k^{(j)}$, we extract the associated clip $\mathcal{C}_k^{(j)}$ and annotate only the central boundary frame $b_k^{(j)}$. The remaining frames within $\mathcal{C}_k^{(j)}$ are left unlabeled, but are used as temporal context during training. This design incurs an annotation cost of $N_q K$ labeled frames per iteration, while exposing the model to $N_q K \ell$ frames of training context per iteration.

\subsection{Training and Active Learning Loop}
\label{sec:training}

After querying, we add the selected clips to the labeled set and update the segmentation model under sparse supervision. For a partially annotated clip $j$, the model processes the clip features $\{\mathbf{x}^{(j)}_t\}_{t\in I^{(j)}_k}$ and outputs per-frame class posteriors $\{\mathbf{p}^{(j)}_t\}_{t\in I^{(j)}_k}$ with $\mathbf{p}^{(j)}_t\in\mathbb{R}^C$. Supervision is applied only to the queried boundary frame(s), while the remaining clip frames are used as unlabeled temporal context. The training objective is the cross-entropy averaged over labeled frames in $\Omega_L$:
\begin{equation}
\mathcal{L}_{\text{total}}=\frac{1}{|\Omega_L|}\sum_{(i,t)\in\Omega_L}\mathrm{CE}\!\left(\mathbf{p}_t^{(i)},y_t^{(i)}\right).
\label{eq:total_loss}
\end{equation}
where each element $(i,t)\in\Omega_L$ indicates that frame $t$ of video $\mathbf{X}^{(i)}$ has an annotation $y_t^{(i)}$.

Alg.~\ref{alg:active_learning} summarizes the complete AL procedure. We initialize $D_L$ with a small randomly sampled subset and define $D_U = D_{train} \setminus D_L$. At each round, the model is trained using Eq.~\eqref{eq:total_loss}, videos are selected according to (Sec.~\ref{sec:video_selection}), and candidate boundaries are scored using Eq.~\eqref{eq:boundary_score}, with the top-K boundaries queried for annotation. Newly labeled boundary indices are added to $\Omega_L$, and the process repeats until the labeling budget $B$ is exhausted or no unlabeled videos remain.

\begin{algorithm}[h]
\caption{Boundary-Centric Clip-Budgeted Active Learning}
\label{alg:active_learning}
\begin{algorithmic}[1]
\STATE \textbf{Input:} $\mathcal{D}_{\text{train}}, \mathcal{D}_{\text{test}}$, rounds $R$, labeled-frame budget $B$, query size $N_q$, clips/video $K$, clip length $\ell$, MC samples $S$
\STATE \textbf{Output:} model $M_\theta$, history $H$
\STATE Initialize $\mathcal{D}_L$ with $n_{\text{init}}$ randomly sampled videos; $\mathcal{D}_U \leftarrow \mathcal{D}_{\text{train}}\setminus\mathcal{D}_L$; initialize labeled indices $\Omega_L$ from $\mathcal{D}_L$
\FOR{$r=1,\dots,R$}
    \STATE Train $M_\theta$ on labeled frames $\Omega_L$ (Eq.~\eqref{eq:total_loss}); evaluate on $\mathcal{D}_{\text{test}}$; append to $H$
    \IF{$|\Omega_L| + N_qK > B$ \OR $|\mathcal{D}_U|=0$} \STATE \textbf{break} \ENDIF
    \STATE \textbf{Stage 1:} Compute $U(\mathbf{X})=\mathrm{mean}_t\,u_t(\mathbf{X})$ for each $X$ in $\mathcal{D}_U$ using $S$ MC-dropout passes Eq.~\eqref{eq:frame_uncertainty}; select $\mathcal{S}_{\text{query}} \leftarrow$ top-$N_q$
    \FOR{each $\mathbf{X}^{(j)}\in \mathcal{S}_{\text{query}}$}
        \STATE \textbf{Stage 2:} Detect candidate boundaries $\mathcal{B}(\mathbf{X}^{(j)})$; score with Eq.~\eqref{eq:boundary_score}; select top-$K$ $\{b_k^{(j)}\}$
        \STATE Query labels at $\{b_k^{(j)}\}$ (use $\ell$-frame clips as unlabeled context); update $\Omega_L \leftarrow \Omega_L \cup \{(j,b_k^{(j)})\}_{k=1}^{K}$
    \ENDFOR
    \STATE $\mathcal{D}_L \leftarrow \mathcal{D}_L \cup \mathcal{S}_{\text{query}}$; $\mathcal{D}_U \leftarrow \mathcal{D}_U \setminus \mathcal{S}_{\text{query}}$
\ENDFOR
\RETURN $M_\theta, H$
\end{algorithmic}
\end{algorithm}

\section{Experiments}
\label{sec:experiments}

\subsection{Datasets}
We evaluate on three standard benchmarks for temporal action segmentation. \textbf{50Salads}~\cite{stein2013combining} contains 50 videos of people preparing salads, annotated with 17 action classes, with an average duration of about 6.4 minutes per video. \textbf{GTEA}~\cite{fathi2011learning} comprises 28 egocentric kitchen videos from 4 subjects, covering 11 action classes; each video is about 1.5 minutes long. \textbf{Breakfast}~\cite{kuehne2014breakfast} includes 1,712 videos of breakfast preparation activities with 48 action classes and roughly 6 action instances per video on average. Together, these datasets span diverse viewpoints, activity complexity, and video lengths, enabling a comprehensive assessment of our approach.

\subsection{Implementation Details}
\label{sec:implementation_details}
For all datasets, we follow common practice and use an I3D~\cite{carreira2017quo} network pretrained on Kinetics to pre-extract per-frame features~\cite{carreira2017quo}. Each frame is represented by a 2048-dimensional I3D feature vector. We report frame-wise accuracy (Acc.), segmental edit score (Edit), and segmental overlap F1 at overlap threshold $k$ ($F1@k$). As the segmentation backbone, we adopt ASFormer~\cite{yi2021asformer} with 10 encoder layers, 3 cascaded decoders, and 64 feature maps. To estimate frame-level uncertainty, we apply Monte Carlo dropout with 10 stochastic forward passes and dropout rate 0.2, and compute predictive entropy. We train using Adam with learning rate $1\times10^{-4}$, weight decay $1\times10^{-5}$, and gradient clipping set to 1. In each active learning iteration, we train the model for 85 epochs.

To ensure a fair comparison, all baselines reported in Tab.~\ref{tab:main_results}, including Su et al.~\cite{su2024two}, use the same ASFormer backbone, training schedule, and
feature extraction pipeline described above. The margin between B-ACT and Su et al.\ in Tab.~\ref{tab:main_results} therefore reflects acquisition strategy differences alone, not architectural advantages.

\textbf{Uncertainty-weighted boundary selection.} Boundary selection follows Sec.~\ref{sec:clip_selection} using the proposed uncertainty-weighted score with $(\alpha,\beta,\gamma)=(0.2,0.3,0.5)$, clip length $\ell=20$, and $K=5$ queried boundaries per video unless otherwise specified.

\textbf{Hyperparameter selection.} The boundary score weights ($\alpha$, $\beta$, $\gamma$) in Eq.~\ref{eq:boundary_score} were selected via grid search on the validation splits and then fixed across all datasets. We searched over $\alpha,\beta \in {0.0, 0.1, \ldots, 0.4}$ and $\gamma \in {0.0, 0.1, \ldots, 0.6}$, subject to the constraint $\alpha + \beta + \gamma = 1$. Each configuration was evaluated on the validation split under the same annotation budget used for testing, and a single configuration was selected based on the combined Edit and Accuracy performance across the validation datasets. Test results were then reported using this fixed set of weights for all datasets. The ablation results in Tab.~\ref{tab:ablation_weights} analyze the contribution of each boundary score component under this shared weighting scheme.


\textbf{Active learning.} We follow a pool-based protocol with sparse, clip-budgeted supervision, tracking annotation cost as a fraction of labeled frames. On GTEA, we initialize with $30\%$ of training videos and run four rounds, each querying $25\%$ of videos. Each queried video is annotated using a single labeled boundary frame per queried clip, while surrounding temporal context of length $\ell$ is used during training. This corresponds to ${\sim}0.4\%$ of frames per video, for a total budget of $0.5\%$ of all training frames. On 50Salads, we use the same four-round protocol with an increased per-video clip budget to reflect longer sequences, initializing with $25\%$ of training videos and querying $25\%$ per round. Each queried video is again annotated with a single labeled boundary frame per queried clip, yielding ${\sim}0.13\%$ of frames per video and a total budget of $0.16\%$ of all training frames.

\subsection{Baselines}
Following \cite{su2024two}, we compare against representative TAS active learning baselines under a unified evaluation pipeline and report the same metrics (F1@\{10,25,50\}, Edit, and frame-wise Accuracy). For clip acquisition, we include Equidistant, Split-Random, Split-Entropy, and Coreset~\cite{sener2017active}. Equidistant samples clips at a fixed temporal stride. Split-Random and Split-Entropy partition each video into four equal temporal segments and select one clip per segment, either uniformly at random or by maximizing predictive entropy. Entropy is estimated via Monte Carlo dropout using 10 stochastic forward passes. Coreset casts selection as a $k$-center problem in the embedding space and is solved with a greedy approximation.

\subsection{Comparison with SOTA}

Table.~\ref{tab:main_results} compares our method with representative baselines and prior state-of-the-art on GTEA, 50Salads, and Breakfast. To ensure a fair comparison, we fix the annotation budget by running all methods for 4 active learning iterations, yielding a total labeled-frame budget of 0.16\% (0.5\% for GTEA). We additionally report an upper bound (\textit{Full}) obtained by training with annotations for all clips.

\begin{table*}[h!]
\centering
\caption{Comparison with state-of-the-art methods and standard baselines on the GTEA, 50Salads, and breakfast datasets. We report F1@\{10,25,50\}, edit, and frame-wise accuracy (acc.) Under the labeling budgets shown. \textbf{bold} indicates the best result in each column.}
\label{tab:main_results}

\setlength{\tabcolsep}{2.2pt}
\renewcommand{\arraystretch}{1.05}
\scriptsize

\resizebox{\textwidth}{!}{%
\begin{tabular}{c|c|ccc|cc|c|ccc|cc|c|ccc|cc}
\noalign{\global\setlength{\arrayrulewidth}{0.8pt}}\hline
\noalign{\global\setlength{\arrayrulewidth}{0.4pt}}
\multirow{2}{*}{ } &
\multicolumn{6}{c|}{\textbf{GTEA}} &
\multicolumn{6}{c|}{\textbf{50Salads}} &
\multicolumn{6}{c}{\textbf{Breakfast}} \\
\cline{2-19}
Method & Budget & \multicolumn{3}{c|}{F1@\{10,25,50\}} & Edit & Acc. & Budget & \multicolumn{3}{c|}{F1@\{10,25,50\}} & Edit & Acc. &  Budget & \multicolumn{3}{c}{F1@\{10,25,50\}} & Edit & Acc. \\

\hline
Random \cite{rana2023hybrid}    & 0.5\%  & 56.1 & 47.9 & 25.5 & 54.1 & 45.2  & 0.16\% & 48.0 & 42.0 & 24.7 & 39.8 & 49.0 & 0.16\% & 61.1 & 55.1 & 39.4 & 56.9 & 61.8 \\
Entropy \cite{rana2023hybrid}   & 0.5\%  & 58.1 & 47.1 & 25.3 & 56.9 & 45.1  & 0.16\% & 39.1 & 34.6 & 16.2 & 35.2 & 45.8 & 0.16\% & 61.9 & 56.8 & 41.0 & 55.8 & 61.8 \\
Equidistant \cite{rana2023hybrid}    & 0.5\%  & 55.3 & 45.1 & 20.7 & 56.3 & 42.7 & 0.16\%  & 44.8 & 38.6 & 27.8 & 36.5 & 51.0 & 0.16\% & 55.6 & 49.2 & 34.0 & 52.0 & 58.5 \\
Coreset \cite{xie2023active}   & 0.5\%  & 50.7  & 41.7  & 23.4  & 50.3  & 43.1 & 0.16\% & 29.1 & 24.9 & 13.8 & 26.1 & 38.4 & 0.16\% & 60.6 & 55.9 & 40.5 & 56.0 & 61.0 \\
Su \textit{et al.} \cite{su2024two} & 0.5\%  & 59.9 & 48.7 & 27.3 & 57.0 & 47.6 & 0.16\% & 55.1 & 49.1 & 32.9 & 45.0 & 57.8 & 0.16\% & 62.8 & 58.1 & 43.5 & 58.6 & 63.5 \\
\hline
B-ACT (Ours)    & 0.5\%  & \textbf{70.8} & \textbf{61.2} & \textbf{42.2} & \textbf{66.6} & \textbf{61.5} & 0.16\% & \textbf{64.7} & \textbf{62.4} & \textbf{52.6} & \textbf{56.7} & \textbf{73.2} & 0.16\% & \textbf{67.5}  & \textbf{61.5} & \textbf{50.9} & \textbf{67.5}& \textbf{70.5} \\
\hline
Full     & 100\%  & 88.6  & 84.4  & 69.2  & 82.5  & 73.4  & 100\% & 82.8 & 80.3 & 67.4  & 73.2 & 82.4 & 100\% & 75.6  & 72.0 & 57.9 & 73.2 & 76.4  \\

\noalign{\global\setlength{\arrayrulewidth}{0.8pt}}\hline
\noalign{\global\setlength{\arrayrulewidth}{0.4pt}}
\end{tabular}
}%
\end{table*}

B-ACT achieves the best performance across all three benchmarks under sparse supervision. On GTEA, it improves $F1@50$ from 27.3 reported by Su \textit{et al.} to 42.2, and also delivers the highest Edit score of 66.6 and the highest frame accuracy of 61.5. On 50Salads, B-ACT establishes a clear margin over all baselines, reaching 64.7, 62.4, and 52.6 on $F1@10$, $F1@25$, and $F1@50$, and improving accuracy from 57.8 to 73.2 compared to \cite{su2024two}. On Breakfast, B-ACT also attains the strongest results, achieving the best $F1@10$, $F1@25$, and $F1@50$ scores of 67.5, 61.5, and 50.9, together with the top Edit score of 67.5 and the highest accuracy of 70.5. Overall, these results suggest that enforcing temporally informative queries is important for TAS, and that our boundary-centric selection consistently improves segmental quality and frame recognition under extremely sparse labeling budgets.

\begin{figure}[h]
    \centering
    \includegraphics[width=0.99\linewidth]{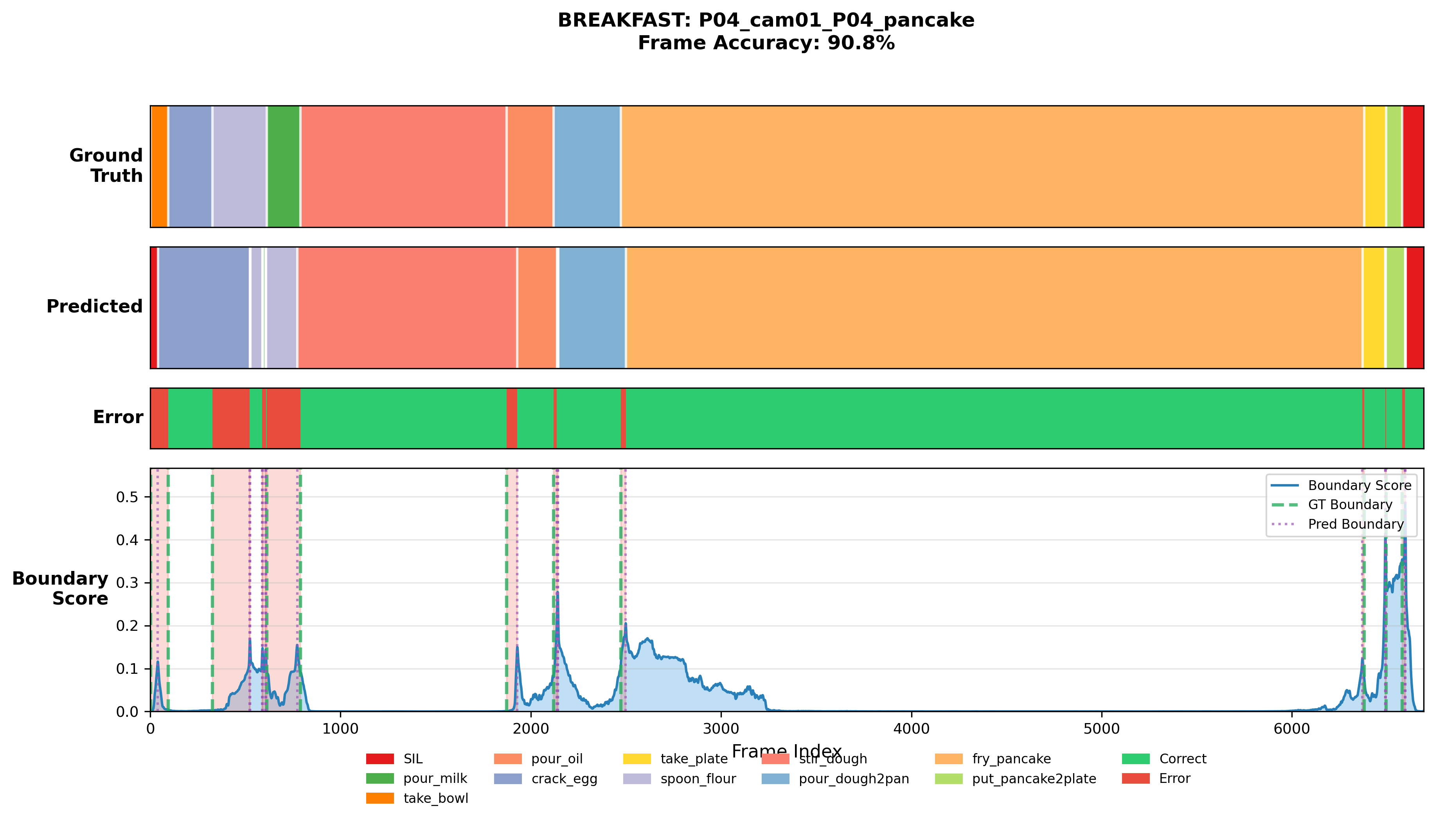}\\[0.5em]
    \includegraphics[width=0.99\linewidth]{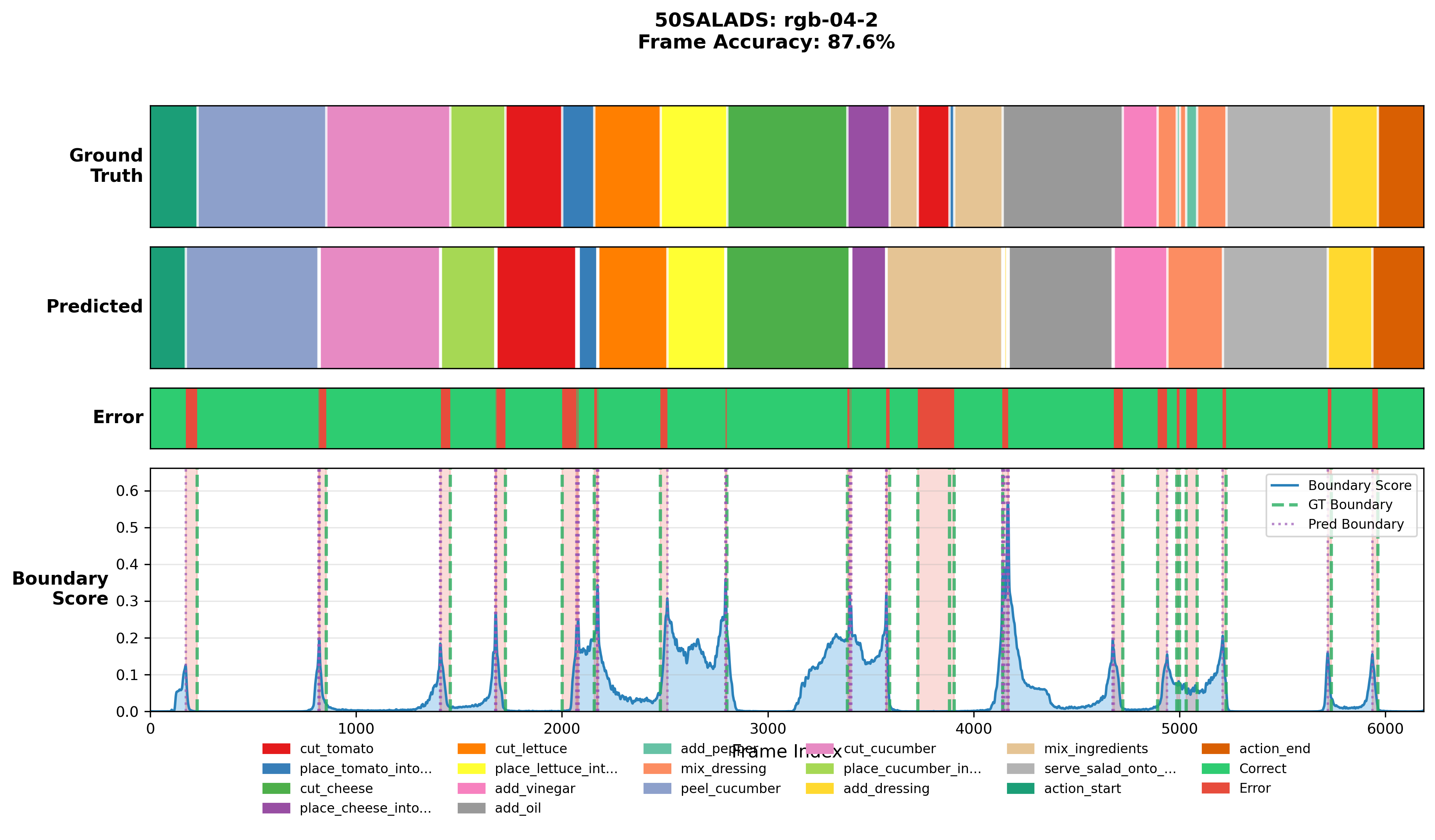}\\[0.5em]
    \includegraphics[width=0.99\linewidth]{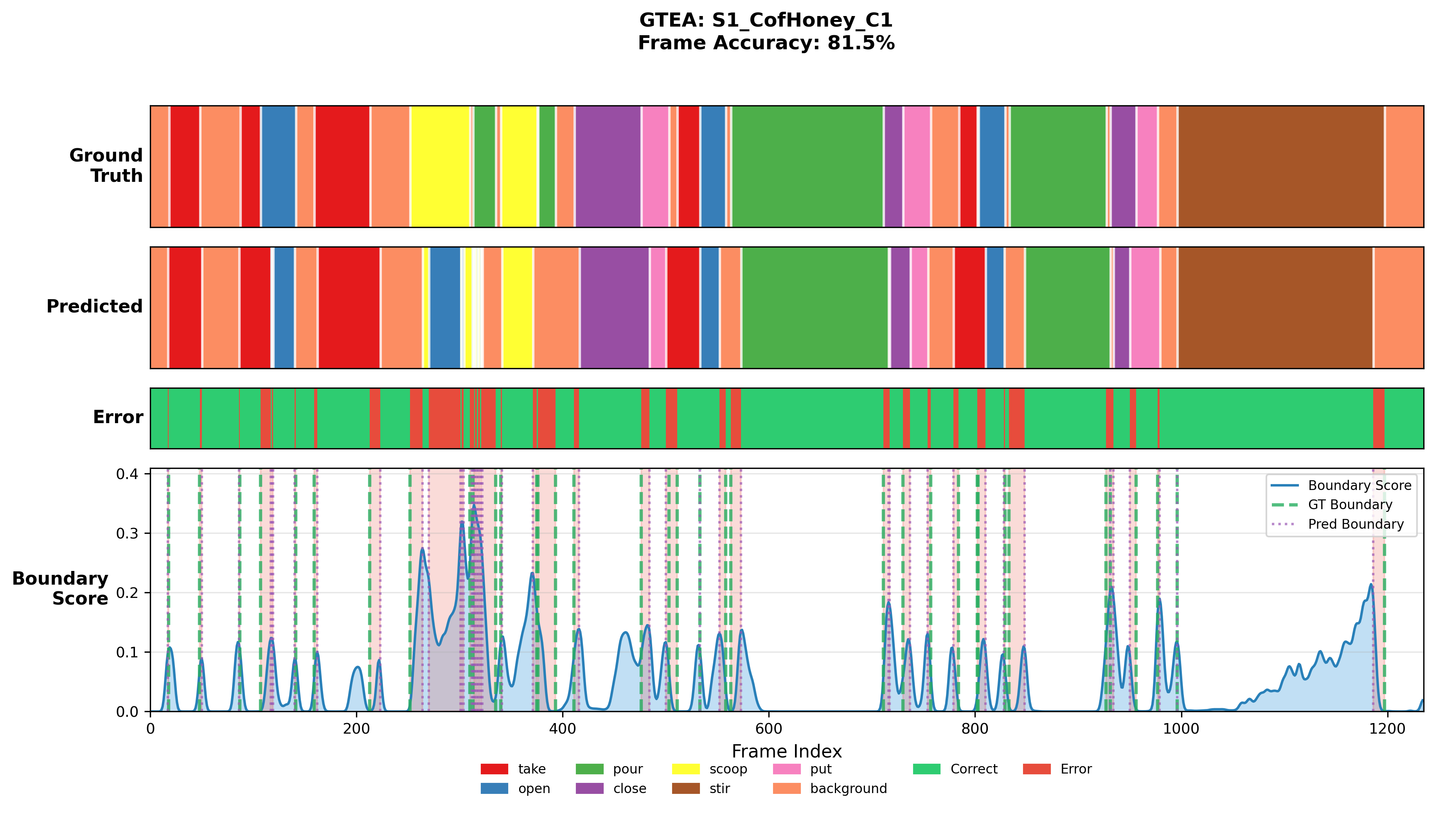}
    \caption{\textbf{Qualitative TAS on sample videos with boundary scores.} Breakfast (top), 50 Salads (middle), and GTEA (bottom) visualize ground truth vs. predicted segments, per-frame errors, and the boundary score with GT/predicted boundaries.}
    \label{fig:actions_detail}

\end{figure}

\textbf{Qualitative Results}. Our qualitative analysis in Fig.~\ref{fig:actions_detail} illustrates the segmentation quality of our model across GTEA, 50Salads, and Breakfast. In the shown examples, the predicted sequences closely match the ground truth, with long action segments recovered with consistent labels and only minor deviations. Most errors are concentrated around action transitions and short-duration actions, indicating that remaining failures are primarily boundary-related rather than sustained confusion over extended intervals. For example, in the shown Breakfast video, the predicted sequence closely matches the main long-duration activity (Acc. 91.8\%), while the GTEA and 50Salads examples preserve the overall temporal ordering and approximate segment durations, with mismatches limited to brief fragments (Acc. 81.7\% and 87.7\% for the shown videos). Overall, the figure suggests that our approach produces stable, temporally coherent segmentations, with residual errors largely confined to ambiguous transition regions.

\subsection{Ablation Studies}

We study how the sampling policy affects performance at two different granularities: (i) \emph{video} selection (which training videos to annotate next), and (ii) \emph{clip} selection (which temporal segments to annotate within the chosen videos). We report results on 50Salads at four labeling budgets (0.06\%, 0.10\%, 0.13\%, 0.16\%) using frame-wise \textbf{Accuracy} and \textbf{Edit Score}.

\begin{figure}[h]
    \centering
    \includegraphics[width=0.9\linewidth]{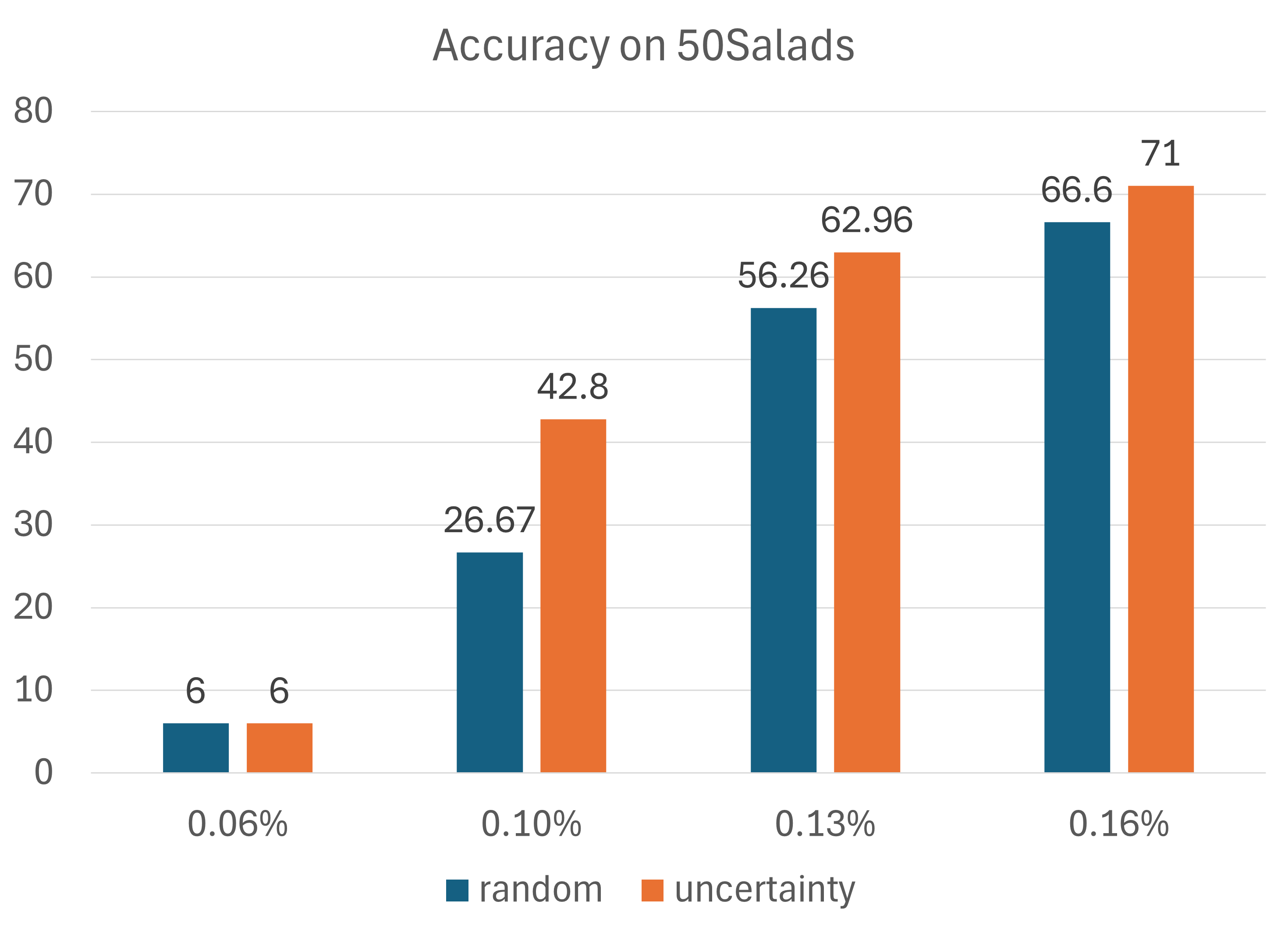}\\[0.5em]
    \includegraphics[width=0.9\linewidth]{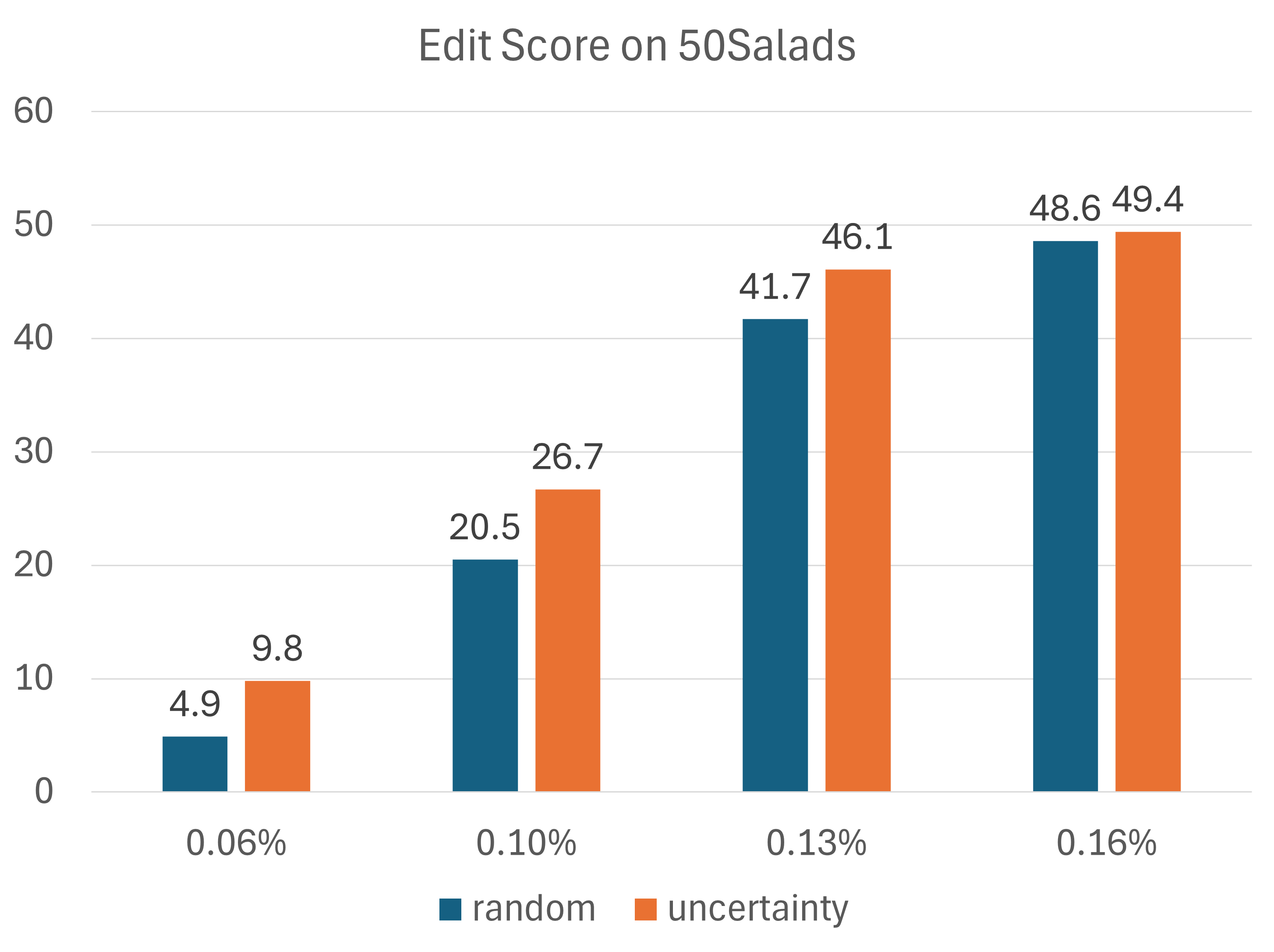}
    \caption{\textbf{Ablation On Video Selection.} Performance as a function of annotation budget when selecting videos uniformly at random versus using uncertainty-based acquisition (higher is better). (Top) Frame-wise accuracy and (bottom) edit score on 50Salads. Uncertainty-based selection yields consistent gains beyond the extremely low-budget regime, indicating that reliable video-level uncertainty estimates improve label efficiency.}
    \label{fig:ablation_video}
\end{figure}

\textbf{Video selection.}
We first ablate the video selection strategy on the 50 Salads dataset by switching between Random and Uncertainty-based sampling while keeping the remainder of the pipeline unchanged, as shown in Fig.~\ref{fig:ablation_video}. Uncertainty-based video selection consistently improves performance once the budget exceeds the extremely low regime. For instance, at 0.10\% budget, uncertainty increases Accuracy from 26.67 to 42.8 and Edit Score from 20.5 to 26.7. The gains persist at higher budgets, reaching 71.0 vs.\ 66.6 Accuracy and 49.4 vs.\ 48.6 Edit Score at 0.16\%. At 0.06\%, both strategies yield the same Accuracy (6), suggesting that this budget is too small for uncertainty estimates to reliably guide video-level acquisition.

This behavior reflects a well-known cold-start limitation of uncertainty-based acquisition \cite{ren2021survey,gal2017deep}. In the earlier rounds, the model is trained on only a handful of partially annotated videos and has not yet learned a discriminative representation of the action space. MC Dropout entropy consequently reflects initialization noise rather than genuine epistemic uncertainty, producing nearly uniform video-level scores across the unlabeled pool and rendering uncertainty-based ranking no more informative than random selection. This is consistent with \cite{su2024two}, whose alignment-based criterion operates on structural features independent of model confidence and therefore remains effective at very low budgets. As supervision accumulates and the model learns class-discriminative boundaries, uncertainty estimates become reliable proxies for annotation value, and uncertainty-based selection yields consistent gains. A warm-start strategy that defers uncertainty-based acquisition until the model is sufficiently trained is a natural extension we leave for future work.

\begin{figure}[h]
    \centering
    \includegraphics[width=0.9\linewidth]{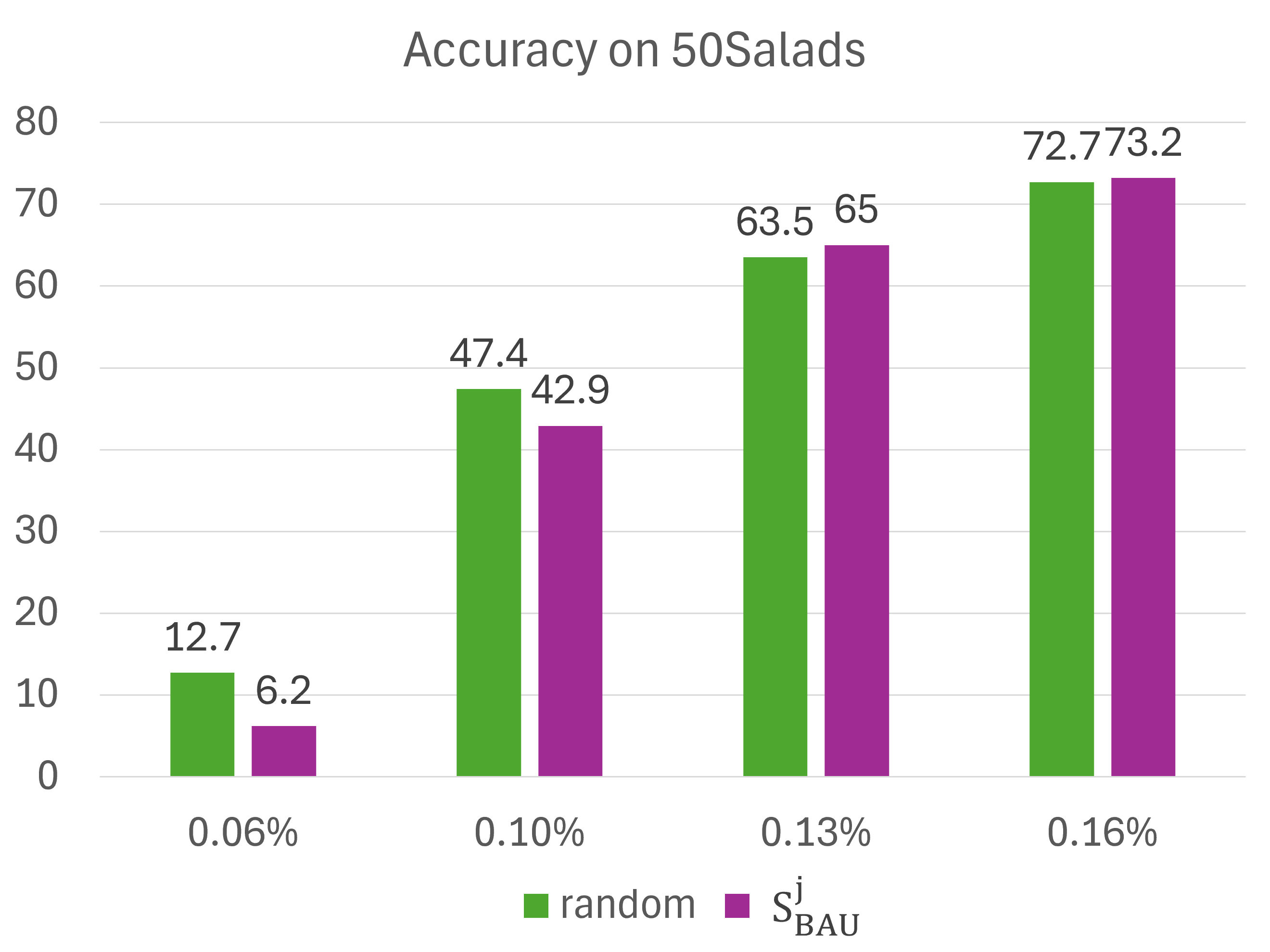}\\[0.5em]
    \includegraphics[width=0.9\linewidth]{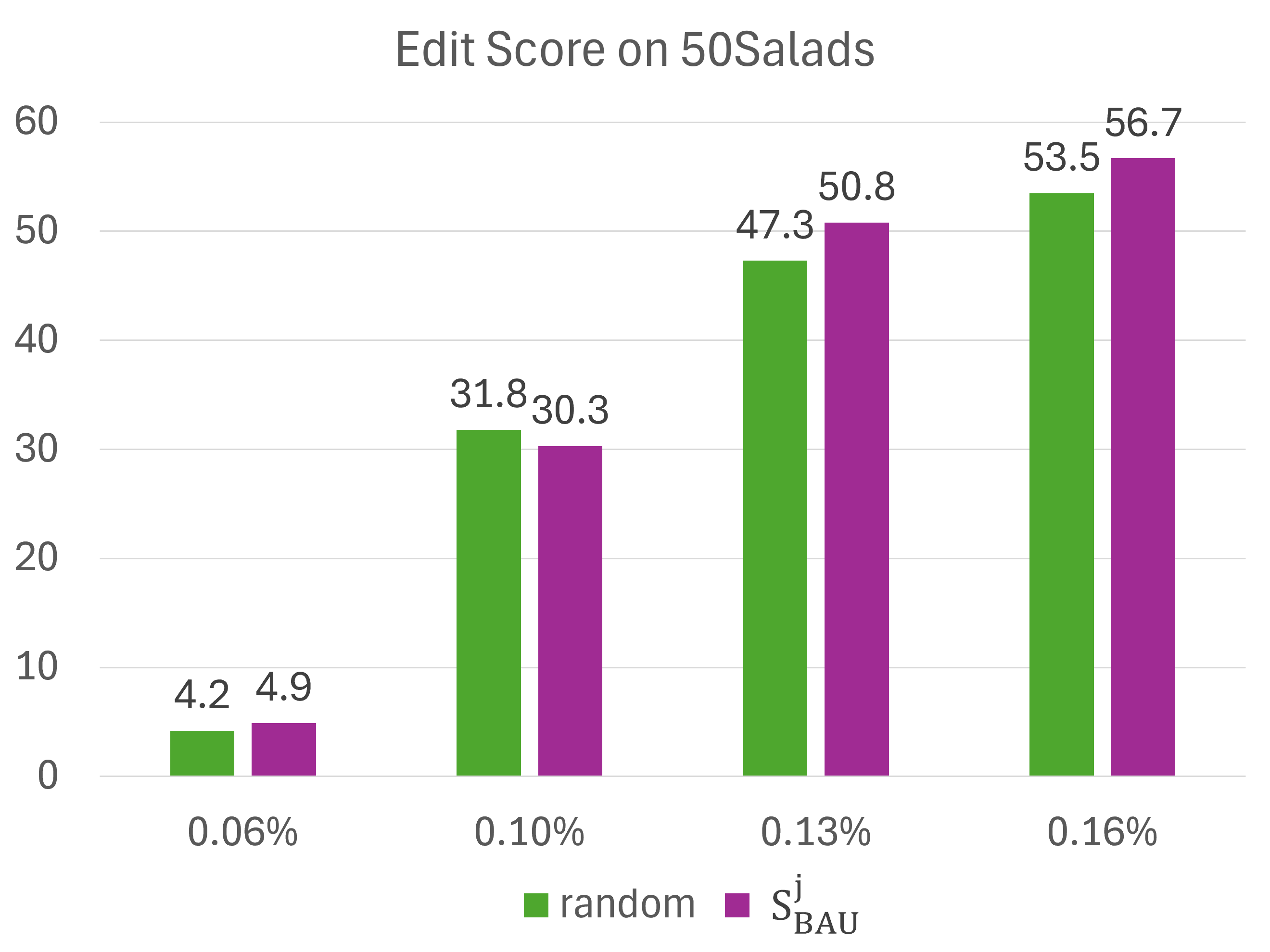}
    \caption{\textbf{Ablation On Clip Selection.} With uncertainty-based \emph{Video Selection} fixed, we compare random clip sampling to the proposed $S_{\mathrm{BAU}}$-based clip selector under varying annotation budgets. (Top) Frame-wise accuracy and (bottom) edit score on 50Salads. $S_{\mathrm{BAU}}$-driven clip selection yields increasing gains as budget grows, while at extremely low budgets boundary-aware scores can be less reliable, making random sampling competitive.}
    \label{fig:ablation_clip}
\end{figure}

\textbf{Clip selection.}
Next, on the 50Salads dataset, we fix the video selection strategy to Uncertainty and isolate the effect of the clip selector by comparing Random clip sampling (green) against the proposed Uncertainty-weighted clip selection (purple). Fig.~\ref{fig:ablation_clip} displays that the uncertainty-weighted clip selector provides clearer improvements as the budget increases. At 0.13\%, it improves Accuracy from 63.5 to 65.0 and Edit Score from 47.3 to 50.8. At 0.16\%, the gains remain consistent, from 72.7 to 73.2 Accuracy and from 53.5 to 56.7 Edit Score. At very low budgets (0.06\% and 0.10\%), random can be competitive or slightly better on Accuracy (e.g., 12.7 vs.\ 6.2 at 0.06\%), indicating that clip-level uncertainty becomes reliable once the model has learned a minimally informative representation. This mirrors the video-level cold-start effect. At very low budgets, model predictions are unreliable, and the predicted boundary set $B_j$ is noisy: the model cannot yet distinguish true action transitions from spurious label flips, causing top-K selection to target uninformative frames. Random sampling is immune to this prediction noise and can incidentally cover true transitions. As the budget grows and predictions stabilize, boundary detection improves and $S_{\mathrm{BAU}}$ consistently outperforms random selection.

\textbf{Boundary score weight ablation.} The results in Table~\ref{tab:ablation_weights} analyze the contribution of each term in the proposed boundary score Eq.~\ref{eq:boundary_score} by enabling $u_b^{\text{local}}$, $g_b$, and $\nabla_b$ individually and in combination, together with their corresponding weights. Using any single component already yields competitive performance, where $\nabla_b$ alone achieves the strongest boundary quality in terms of edit score (61.5) and improves F1 across all thresholds (65.2/57.9/42.7), indicating that boundary-gradient cues are particularly informative for temporally coherent segmentation. In contrast, $g_b$ alone attains the highest accuracy among the single-term settings (62.3), suggesting that global boundary evidence is beneficial for frame-level correctness. However, naive uniform combinations degrade performance, for instance, equal weights of 0.33 reduce accuracy to 57.6 and edit to 52.3, highlighting that these cues are not equally reliable and can interfere when aggregated without calibration. By re-balancing the contributions, the weighted fusion $w_{u_b^{\text{local}}}=0.2$, $w_{g_b}=0.3$, $w_{\nabla_b}=0.5$ provides the best overall trade-off, achieving competitive accuracy (61.5) and the best F1 scores (70.8/61.2/42.2), while maintaining a high edit score (66.6). The dominant weight on $\gamma$ reflects that temporal gradient is the strongest contributor to the boundary quality in our ablations (Table II), while the contributions of local uncertainty and ambiguity provide complementary regularization that is beneficial once the model has learned a minimally informative representation.

It is worth examining whether local uncertainty and ambiguity provide complementary signal beyond $\nabla_b$ alone. From Tab.~\ref{tab:ablation_weights}, $\nabla_b$ alone achieves a competitive single-term edit score (61.5) and the best F1@50 (42.7), but falls short of the full combination on edit score (61.5 vs.\ 66.6) and F1@10 (65.2 vs.\ 70.8), indicating that $u_b^{\text{local}}$ and $g_b$ contribute complementary signal. This indicates that temporal gradient captures boundary sharpness — a necessary but not sufficient condition for annotation value. Local uncertainty $u_b^{\text{local}}$ adds robustness in regions where the gradient is large but the model is already confident, filtering spurious sharp transitions that do not represent genuine action changes. The confidence gap $g_b$ separately targets boundaries where competing classes are nearly tied — a distinct failure mode from distributional sharpness that global entropy may understate. The three terms are therefore not redundant: each filters a qualitatively different class of uninformative boundaries, and their weighted combination achieves the best edit score (66.6) and F1@10 (70.8) in Tab.~\ref{tab:ablation_weights}, outperforming pairwise variant on these metrics.

\begin{table}[h]
\centering
\caption{Ablation study on component weighting for boundary-aware clip selection on GTEA. We evaluate individual components (rows 1--3), pairwise combinations (rows 4--6), uniform weighting (row 7), and optimized weighting (row 8). The full model with weights (0.2, 0.3, 0.5) achieves the best overall performance.}
\label{tab:ablation_weights}
\setlength{\tabcolsep}{3.5pt}
\renewcommand{\arraystretch}{1.05}
    \begin{tabular}{ccc| ccc| cc| ccc}
    \hline
    $u_b^{\text{local}}$ & $g_b$ & $\nabla_b$ &
    $w_{u_b^{\text{local}}}$ & $w_{g_b}$ & $w_{\nabla_b}$ &
    Edit & Acc. & \multicolumn{3}{c}{F1@\{10,25,50\}} \\
    \hline
    $\checkmark$ & $\times$      & $\times$      & 1.0  & 0.0  & 0.0  & 54.2 & 60.1 & 61.1 & 56.9 & 39.5 \\
    $\times$      & $\checkmark$ & $\times$      & 0.0  & 1.0  & 0.0  & 56.1 & 62.3 & 60.4 & 53.9 & 37.3 \\
    $\times$      & $\times$      & $\checkmark$ & 0.0  & 0.0  & 1.0  & 61.5 & 61.8 & 65.2 & 57.9 & \textbf{42.7} \\
    $\checkmark$ & $\checkmark$ & $\times$      & 0.6  & 0.4  & 0.0  & 55.0 & 58.5 & 64.7 & 57.4 & 38.1 \\
    $\checkmark$ & $\times$      & $\checkmark$ & 0.4  & 0.0  & 0.6  & 56.2 & 63.6 & 61.8 & 57.1 & 42.6 \\
    $\times$      & $\checkmark$ & $\checkmark$ & 0.0  & 0.4  & 0.6  & 64.6 & \textbf{66.0} & 63.8 & 57.9 & 42.4 \\
    $\checkmark$ & $\checkmark$ & $\checkmark$ & 0.33 & 0.33 & 0.33 & 52.3 & 57.6 & 60.3 & 53.5 & 38.5 \\
    \hline
    $\checkmark$ & $\checkmark$ & $\checkmark$ & 0.2  & 0.3  & 0.5  & \textbf{66.6} & 61.5 & \textbf{70.8} & \textbf{61.2} & 42.2 \\
    \hline
\end{tabular}
\end{table}

\begin{figure*}[tbp]
    \centering
    \includegraphics[width=0.8\textwidth]{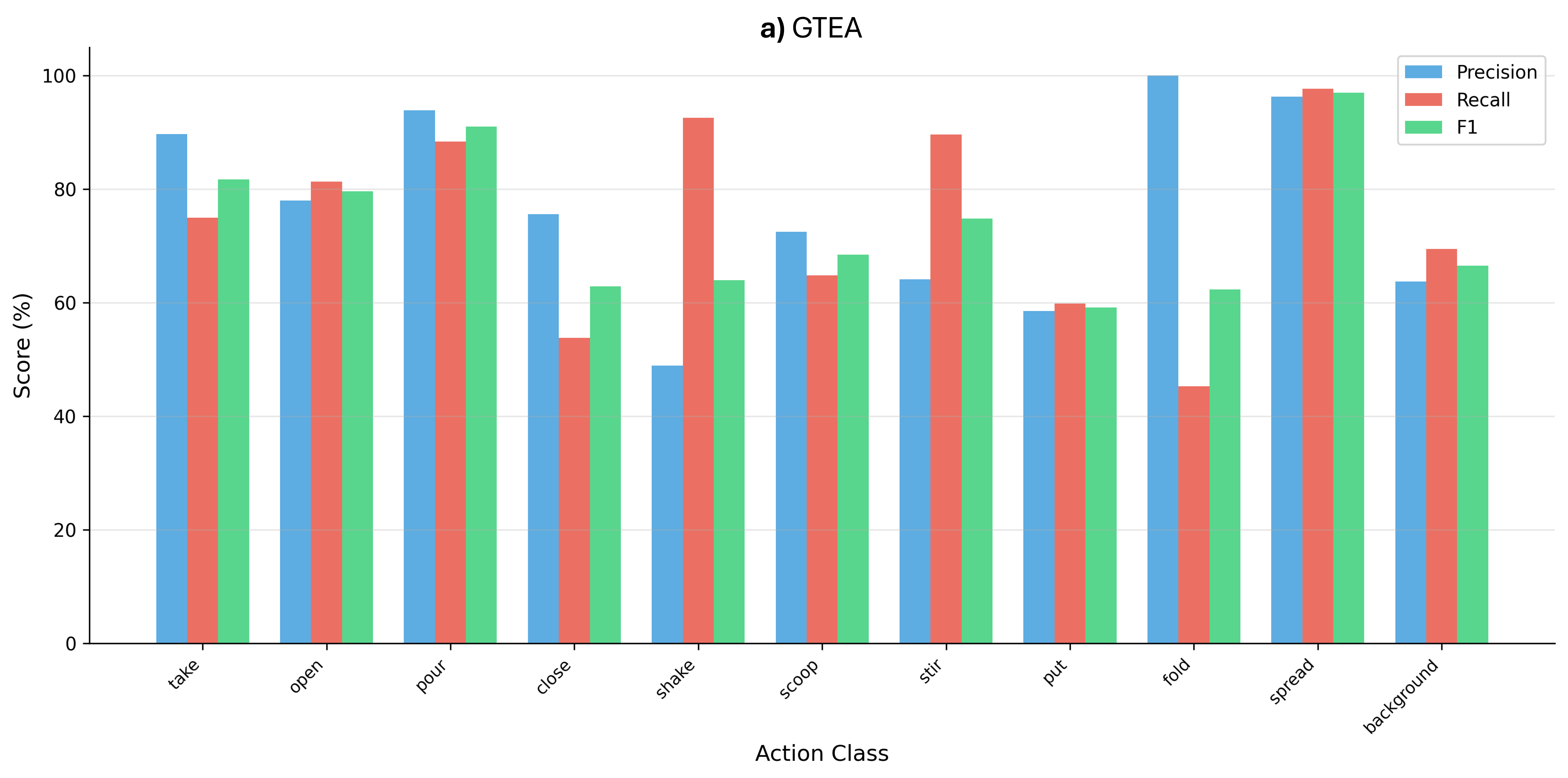}
    \caption{\textbf{Per-class performance on GTEA.} Precision, recall, and F1 for each action class. Frequent, manipulation-heavy actions (e.g., \textit{take}, \textit{put}) score highest, while short transitional actions show the largest F1 drop.}
    \label{fig:perclass_gtea}
\end{figure*}

\begin{figure*}[tbp]
    \centering
    \includegraphics[width=0.8\textwidth]{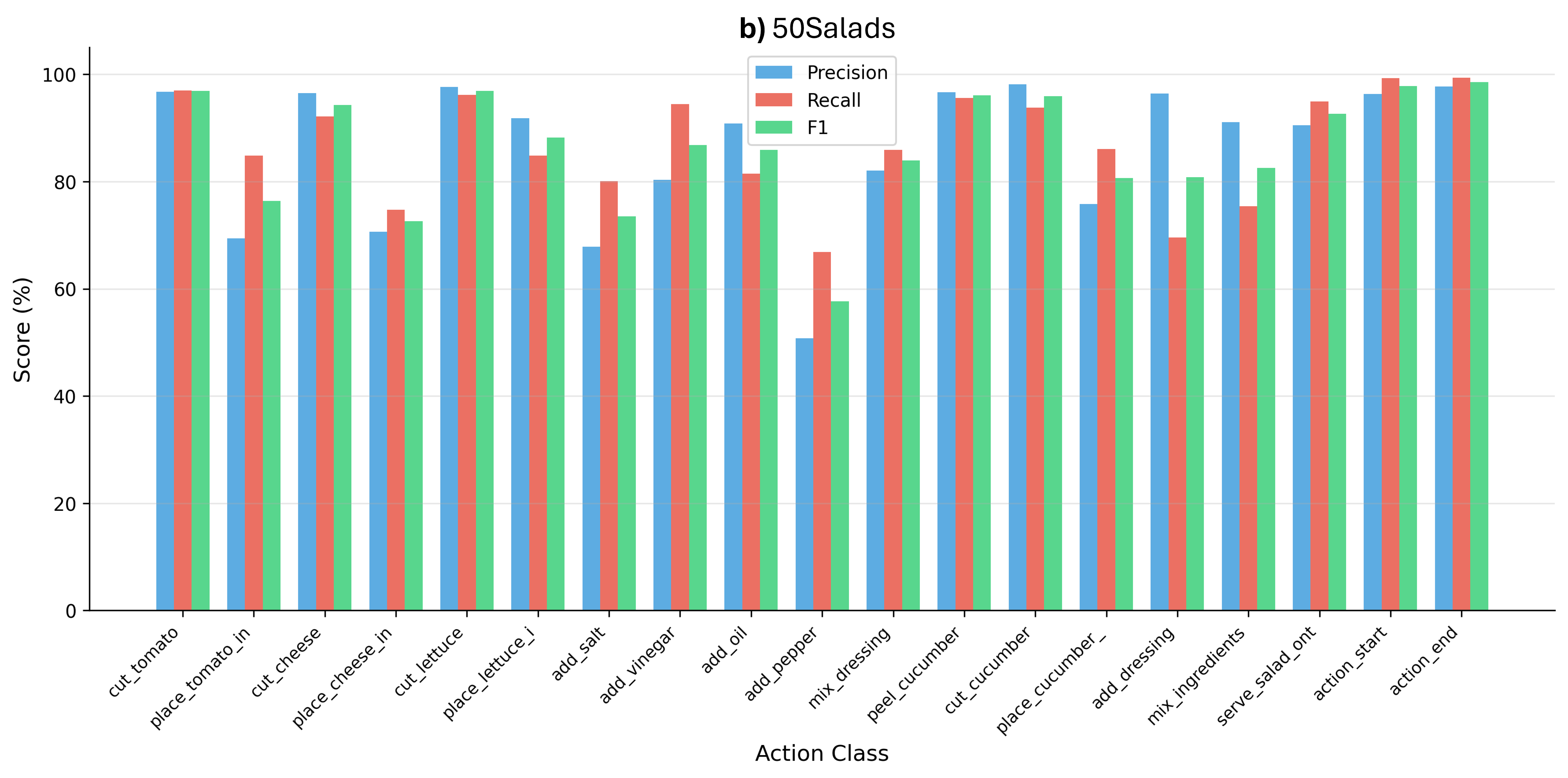}
    \caption{\textbf{Per-class performance on 50Salads.} Precision, recall, and F1 for each action class. Cutting and mixing actions are recovered reliably; visually similar preparation steps (e.g., \textit{add\_oil} vs.\ \textit{add\_vinegar}) account for most of the residual error.}
    \label{fig:perclass_50salads}
\end{figure*}

\begin{figure*}[tbp]
    \centering
    \includegraphics[width=0.8\textwidth]{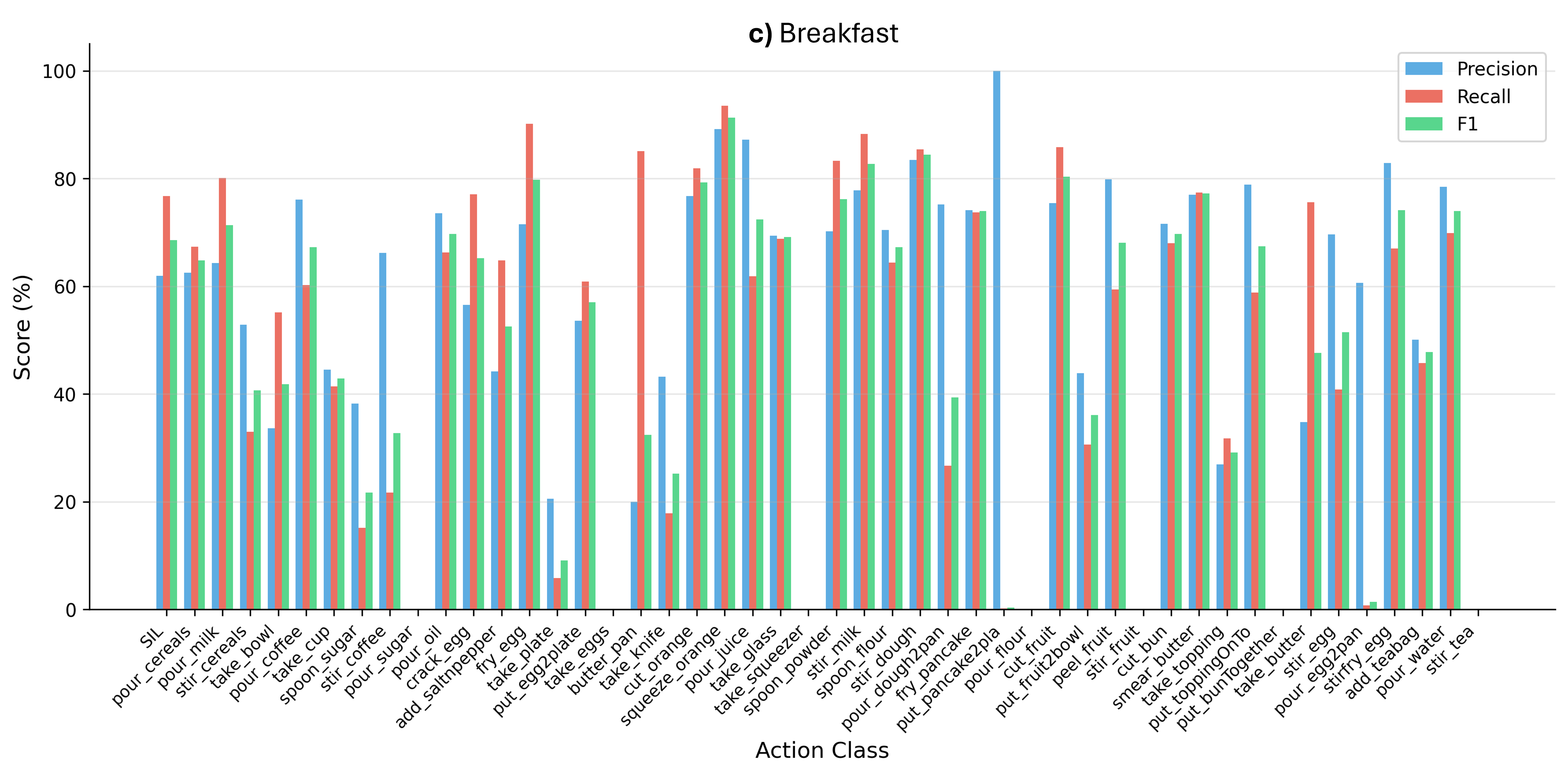}
    \caption{\textbf{Per-class performance on Breakfast.} Precision, recall, and F1 for each action class. The long-tail distribution is visible: head classes saturate near ceiling, while rare and brief actions remain the dominant source of boundary error under sparse supervision.}
    \label{fig:perclass_breakfast}
\end{figure*}

\textbf{Effect of boundary context window length.} We ablate the temporal context size used to form boundary-centered clips by varying the clip length (number of frames sampled around the boundary) and evaluate on GTEA, as shown in Tab.~\ref{tab:clip_length}. Using overly long clips (e.g., 40--50 frames) consistently degrades performance, likely because the clip spans multiple sub-actions and introduces irrelevant context that blurs the transition signal, leading to noisier boundary representations and less effective boundary supervision. In contrast, very short clips provide insufficient temporal evidence to disambiguate the action change and capture pre-/post-boundary dynamics, which reduces Edit and F1, particularly at stricter overlap thresholds. We find that a moderate window offers the best trade-off: a clip length of 20 frames yields the strongest results across all metrics, improving both segmentation quality (Edit) and frame-wise accuracy, while delivering the highest F1 at \{10,25,50\}.

\begin{table}[h]
\centering
\normalsize
\caption{Effect of boundary clip length on GTEA. Ablation over the number of frames sampled around each boundary when forming boundary-centered clips. We report edit, frame-wise accuracy (acc), and F1@\{10,25,50\}.}
\label{tab:clip_length}
\setlength{\tabcolsep}{5pt}
\renewcommand{\arraystretch}{1.1}
\begin{tabular}{c| cc| ccc}
\hline
Clip Length & Edit & Acc. & \multicolumn{3}{c}{F1@\{10,25,50\}} \\
\hline
50 & 40.2 & 43.1 & 37.3 & 29.5 & 15.6 \\
40 & 50.4 & 57.9 & 58.5 & 50.3 & 37.4 \\
30 & 55.3 & \textbf{61.8} & 58.7 & 51.0 & 37.0 \\
\textbf{20} & \textbf{66.6} & 61.5 & \textbf{70.8} & \textbf{61.2} & \textbf{42.2} \\
10 & 58.4 & 61.6 & 63.3 & 53.9 & 33.2 \\
0  & 52.8 & 55.9 & 54.8 & 47.5 & 36.7 \\
\hline
\end{tabular}
\end{table}

\textbf{Per-class performance analysis.}
In Breakfast, the per-class breakdown shows substantial variance across the large action set: many frequent, visually distinctive actions achieve strong and balanced precision and recall, while several fine-grained, visually similar actions (for example, multiple “pour” and “stir” variants) exhibit noticeably lower F1. Errors tend to manifest as precision–recall imbalance, suggesting confusion among closely related action definitions and sensitivity to short transition regions for rare or brief classes.

For 50Salads, performance is generally strong and consistent across most classes, with many actions achieving high precision and recall simultaneously (for example, cutting and peeling actions), indicating stable recognition of the core preparation steps. The main degradations appear in a small subset of classes where either recall drops (missed instances) or precision drops (confusions with neighboring steps), which is typical for actions that are temporally interleaved or visually subtle (for example, seasoning or mixing-related classes). Boundary-related meta-classes such as action start and end are predicted very reliably.

In GTEA, the class-wise results are mostly high for dominant manipulation actions (for example, take, open, pour, spread), but several classes show clear precision–recall asymmetry, indicating systematic confusions. In particular, some actions have very high precision but lower recall (missed occurrences), while others show the opposite pattern (over-prediction), consistent with ambiguities around transitions and short-duration segments. Overall, the remaining errors are concentrated in a few ambiguous classes rather than being uniformly distributed across the label set.

\textbf{Acquisition Function Ablation.} We ablate the acquisition function used for video-level uncertainty estimation in Stage~1. In addition to predictive Entropy~\cite{shannon1948mathematical}, we consider BALD~\cite{houlsby2011bayesian}, Power-BALD~\cite{kirsch2021powerBALD}, Jensen--Shannon divergence (JSD)~\cite{lin1991jsd}, and Variation Ratio~\cite{freeman1965variation}. All acquisition scores are computed from the same \(S=10\) Monte Carlo Dropout forward passes~\cite{gal2016dropout} used throughout the paper. Stage~2 boundary selection and all remaining hyperparameters are kept fixed. Results on GTEA at AL round~4 are reported in Table~\ref{tab:acq_ablation}.

Entropy gives the strongest segmental performance, achieving the best Edit score (66.58) and the best F1@\{10,25\} (70.75 / 61.22). We believe this is because Stage~1 requires a \emph{video-level} acquisition signal obtained by aggregating frame-wise uncertainty across the full sequence. Predictive entropy measures the overall spread of the mean predictive distribution and therefore provides a stable summary of how broadly uncertain the model is over a video. This matches our objective well: videos are valuable when they contain many uncertain transition regions, not merely a few isolated frames with high disagreement.

By contrast, BALD, Power-BALD, and JSD place greater emphasis on disagreement structure across Monte Carlo samples. While this can improve frame-wise discrimination, it appears less well aligned with the segmental goals of TAS after video-level pooling. In particular, Power-BALD attains the highest test accuracy (64.42) and slightly improves F1@50 (43.92), but still underperforms Entropy on Edit and low-threshold F1, indicating that its selected videos do not improve boundary localization as effectively. BALD and JSD show the same pattern more clearly, with moderate accuracy but weaker segmental performance. Variation Ratio performs worst overall, likely because it collapses the predictive distribution too aggressively and discards information about class uncertainty that is useful for dense temporal prediction.

Overall, these results suggest that for Stage~1 of B-ACT, a good acquisition function should provide a robust sequence-level summary of predictive uncertainty rather than a highly selective disagreement signal. Predictive Entropy best matches this requirement, and we therefore use it as the default acquisition function in all main experiments.

\begin{table}[t]
\centering
\caption[Acquisition Function Ablation]{Acquisition Function Ablation on GTEA (round~4, budget 0.5\%). Stage~2 boundary selection and all other hyperparameters are fixed across rows. Bold indicates the best result in each column.}
\label{tab:acq_ablation}
\setlength{\tabcolsep}{3pt}
\resizebox{\columnwidth}{!}{%
\begin{tabular}{l | c | c | c | c | c}
Acq. Function & Acc. & Edit & F1@10 & F1@25 & F1@50 \\
\hline
BALD~\cite{houlsby2011bayesian} & 0.619 & 57.26 & 61.82 & 58.18 & 36.97 \\
Power-BALD~\cite{kirsch2021powerBALD} & \textbf{0.644} & 59.70 & 67.06 & 59.35 & \textbf{43.92} \\
JSD~\cite{lin1991jsd} & 0.627 & 55.17 & 64.12 & 57.65 & 43.53 \\
Variation Ratio~\cite{freeman1965variation} & 0.560 & 54.25 & 56.02 & 48.19 & 33.73 \\
Entropy~\cite{shannon1948mathematical} & 0.615 & \textbf{66.58} & \textbf{70.75} & \textbf{61.22} & 42.18 \\
\end{tabular}%
}
\end{table}

\FloatBarrier
\section{Conclusion}

In this paper we presented B-ACT, a boundary-centric active learning framework for temporal action segmentation that allocates supervision to action transitions. B-ACT combines uncertainty-based video selection with a boundary score that fuses local uncertainty, class ambiguity, and temporal prediction dynamics, querying only a few boundary frames per video while training with boundary-centered context. Experiments on GTEA, 50Salads, and Breakfast demonstrate strong label efficiency and consistent gains over active learning baselines under extremely sparse budgets.

\textbf{Limitations}. A limitation of the current framework is the reduced reliability of uncertainty-based acquisition during the earliest active learning rounds, when only limited supervision is available and uncertainty estimates remain poorly calibrated. As observed in our ablation studies, this cold-start effect can reduce the effectiveness of uncertainty-guided video selection at extremely sparse budgets. Future work will investigate warm-start and hybrid acquisition strategies to improve early-round query efficiency.

\FloatBarrier

\bibliography{main}
\vfill

\end{document}